%% file: main.tex
\documentclass{article}

\PassOptionsToPackage{numbers, compress}{natbib}
\usepackage[preprint]{neurips_2023}
\usepackage[utf8]{inputenc} % allow utf-8 input
\usepackage[T1]{fontenc}    % use 8-bit T1 fonts
\usepackage{hyperref}       % hyperlinks
\usepackage{url}            % simple URL typesetting
\usepackage{booktabs}       % professional-quality tables
\usepackage{amsfonts}       % blackboard math symbols
\usepackage{nicefrac}       % compact symbols for 1/2, etc.
\usepackage{microtype}      % microtypography
\usepackage[table,dvipsnames]{xcolor}
\usepackage{enumitem}
\usepackage{graphicx}
\usepackage{multirow}
\usepackage{pifont}
\usepackage{tcolorbox}
\usepackage{adjustbox}
\usepackage{amsmath,amssymb}
\usepackage{caption}
\usepackage{subcaption}
\usepackage{latexsym}
\usepackage{gensymb}
\usepackage{makecell}
\usepackage{marvosym}
\usepackage{tikz}
\usepackage{pgfplots}
\usepackage{xspace}
\usepackage{wrapfig}
\usepackage{cinzel}
\usepackage[T1]{fontenc} 
\newcommand{\clipvit}{CLIP-ViT\xspace}

\newcommand{\ptvit}{pretrained-ViT\xspace}

\newcommand{\vit}{ViT\xspace}

\newcommand{\methodname}{{\scshape Vit-lens}\xspace}
\usepackage{mdframed}

\newcommand{\upperf}[1]{\textcolor{teal}{(+#1)}}
\newcommand{\celldouble}[2]{\begin{tabular}[c]{@{}c@{}} \textbf{#1} \\ \textbf{#2} \end{tabular}}

\definecolor{mypurple}{RGB}{200,192,248}
\definecolor{mypurpledeep}{RGB}{142,126,240}
\definecolor{mygreen}{RGB}{117,170,156}
\definecolor{myyellow}{RGB}{255,192,0}
\definecolor{myblue}{RGB}{57,143,255}
\definecolor{mygrey}{RGB}{231,230,230}
\definecolor{codey}{RGB}{220,220,170}
\definecolor{coder}{RGB}{206,145,120}
\definecolor{codeb}{RGB}{156,220,254}
\definecolor{codenum}{RGB}{204,204,204}
\definecolor{ADark}{rgb}{0,0.3,0.8}
\definecolor{BDark}{rgb}{.5,.0,.5}
\definecolor{CDark}{rgb}{0,.5,0}
\definecolor{DDark}{rgb}{0.11764705882352941, 0.5647058823529412, 1.0}
\definecolor{EDark}{rgb}{0.8823529411,0.63725490196,0.0156862745}
\definecolor{FDark}{rgb}{0.6235294117647059, 0.27058823529411763, 0.4627450980392157}

\newcommand{\dsA}{\textcolor{myyellow}{$\blacktriangleright$}}
\newcommand{\dsB}{\textcolor{mygreen}{$\blacktriangleright$}}
\newcommand{\dsC}{\textcolor{mypurpledeep}{$\blacktriangleright$}}

\title{\includegraphics[scale=0.040, bb=-100 80 300 34]{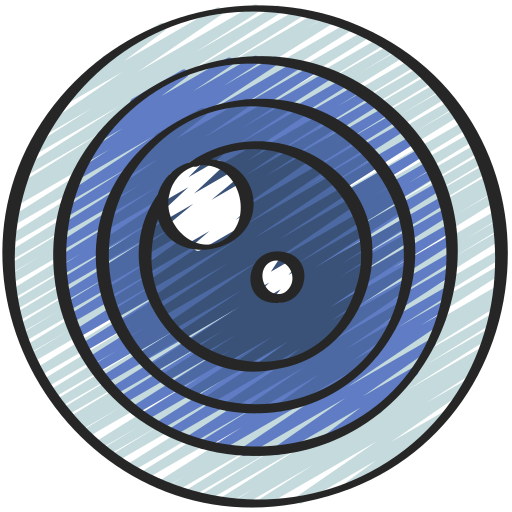}
\methodname: Initiating Omni-Modal Exploration through 3D Insights}

\definecolor{pearThree}{HTML}{E74C3C}
\definecolor{pearDark}{HTML}{2980B9}
\definecolor{pearDarker}{HTML}{1D2DEC}
\definecolor{lavenderblue}{rgb}{0.8, 0.8, 1.0}
\definecolor{lavenderweb}{rgb}{0.9, 0.9, 0.98}

\newcommand{\cmark}{\textcolor{mygreen}{\ding{51}}} % ✔
\newcommand{\xmark}{\textcolor{BrickRed}{\ding{55}}} % ✘

\hypersetup{
 colorlinks,
 citecolor=pearDark,
 linkcolor=pearThree,
    breaklinks=true,
 urlcolor=pearDarker}
 
\author{Weixian Lei\textsuperscript{\rm 1,2,3}\quad
     Yixiao Ge\textsuperscript{\rm 2}\textsuperscript{\dag}\quad
     Jianfeng Zhang\textsuperscript{\rm 3}\quad
     Dylan Sun\textsuperscript{\rm 2}\quad
     Kun Yi\textsuperscript{\rm 2}\quad \\
     \textbf{Ying Shan\textsuperscript{\rm 2}\quad
     Mike Zheng Shou\textsuperscript{\rm 1,3}\textsuperscript{\dag}} \\~\\ 
    \textsuperscript{\rm 1}Show Lab,\textsuperscript{\rm 3}National University of Singapore \quad
    \textsuperscript{\rm 2}ARC Lab, Tencent PCG  \\~\\
    \url{https://github.com/TencentARC/ViT-Lens}
}

\begin{document}

\maketitle
\renewcommand{\thefootnote}{\dag}
\footnotetext{Corresponding authors.}

\begin{abstract} 
Though the success of CLIP-based training recipes in vision-language models, their scalability to more modalities (e.g., 3D, audio, etc.) is limited to large-scale data, which is expensive or even inapplicable for rare modalities.
% given the increasing computational overhead and the lack of naturally paired data.
In this paper, we present \methodname that facilitates efficient omni-modal representation learning by perceiving novel modalities with a \ptvit and aligning to a pre-defined space.
% enhances omni-modal understanding by leveraging a \ptvit to perceive diverse modalities. 
% Recent advances in Vision Language Pretraining have demonstrated remarkable effectiveness in zero-shot understanding and representation learning by aligning multimodal features from images and language descriptions. However, these models are limited in their direct comprehension of sensory modalities beyond images.
Specifically, the modality-specific lens is tuned to project multimodal signals to the shared embedding space, which are then processed by a strong ViT that carries pre-trained image knowledge.
% The encoded multimodal representations are optimized toward the objective of aligning with the multimodal space that is pre-defined by off-the-shelf foundation models.
% The well-trained lens equipped with ViT backbone serves as one of the foundation models that supervise the optimization of the next modality.
% The encoded multimodal representations are optimized to align with a pre-defined multimodal space, derived from off-the-shelf foundation models, which include the initial CLIP models and pre-trained lens with a ViT backbone.
The encoded multimodal representations are optimized toward aligning with the modal-independent space, pre-defined by off-the-shelf foundation models. A well-trained lens with a ViT backbone has the potential to serve as one of these foundation models, supervising the learning of subsequent modalities.
\methodname provides a unified solution for representation learning of increasing modalities with two appealing benefits:
(i) Exploiting the \ptvit across tasks and domains effectively with efficient data regime;
(ii) Emergent downstream capabilities of novel modalities are demonstrated due to the modality alignment space.
% The recent advancements in Vision Foundation Models and the strides made in Vision Language Pretraining have demonstrated their impressive efficacy across various understanding tasks and representation learning abilities.
% While existing methods attempt omni-modal understanding through separate architectures or large-scale data collection, these approaches lack scalability to accommodate diverse modalities. 
% Inspired by the exceptional capabilities and transferability of the \ptvits, \methodname conceptualizes a \ptvit as a multi-modal sensor with the ability to comprehend various modalities. Instead of solely aligning to pre-established feature spaces, our method encodes out-of-image modalities through a \ptvit to optimize knowledge utilization.
% \methodname offers several advantages: (1) Model Unification, employing a shared \vit for various modalities, facilitating scalability and convergence. (2) Data-Efficient, effectively utilizing the \ptvit across tasks and domains without the need for large-scale data. (3) Emergent Ability, enabling Large Language Models to seamlessly comprehend novel modalities without specific tuning.
We evaluate \methodname in the context of 3D as an initial verification.
% We evaluate \methodname in the context of 3D shape understanding. 
In zero-shot 3D classification, \methodname achieves substantial improvements over previous state-of-the-art, showing \textbf{52.0\%} accuracy on Objaverse-LVIS, \textbf{87.4\%} on ModelNet40, and \textbf{60.6\%} on ScanObjectNN. 
Furthermore, we enable zero-shot 3D question-answering by simply integrating the trained 3D lens into the InstructBLIP model without any adaptation.
% Additionally, when integrated into InstructBLIP, our method enables the Large Language Model to understand 3D shapes in a zero-shot manner.
% We hope that our work will serve as a source of inspiration, encouraging researchers to delve deeper into the realm of omni-modal understanding.
% We hope that our work will inspire researchers to delve deeper into omni-modal representation learning.
We will release the results of \methodname on more modalities in the near future.
% , and hope that our work could inspire the community to delve deeper into omni-modal representation learning.

\end{abstract}

\section{Introduction}

\begin{figure}[th]
  \centering
  \includegraphics[width=\textwidth]{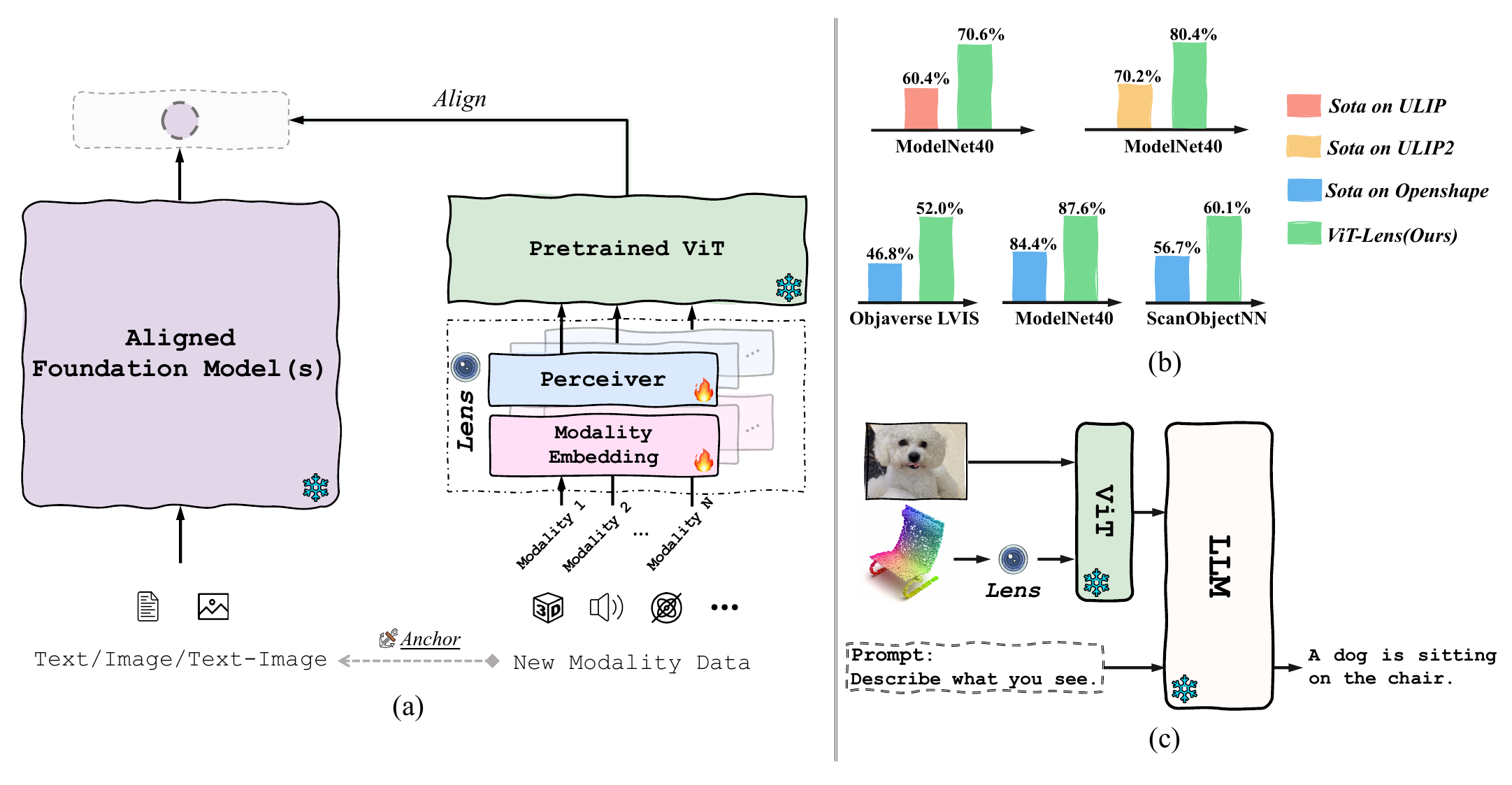}
  \caption{\textbf{(a) Illustration of \methodname.} \methodname extends the capabilities of a \ptvit to perceive and comprehend diverse modalities beyond 2D images. It achieves this by firstly employing Modality Embedding and the Perceiver architecture~\cite{jaegle2021perceiver} to map modality-specific data into the \ptvit input space. Then the encoded output of \vit is aligned with the feature extracted from the data's anchor text/image/text-image, through an off-the-shelf foundation model.
  This novel approach enables a \ptvit to integrate and understand diverse modalities beyond images while leveraging its knowledge from the pretraining to better comprehend and interpret these modalities.
  \textbf{(b) Zero-shot 3D classification.} Our \methodname outperforms the state-of-the-art methods on zero-shot 3D classification when pretrained on datasets introduced by ULIP~\cite{xue2023ulip}, ULIP2~\cite{xue2023ulip2} and OpenShape~\cite{liu2023openshape} respectively.
  \textbf{(c) Emergent Downstream Abilities.} By incorporating new modalities into the \vit of an off-the-shelf MLLM, \methodname empowers the LLM to understand novel modalities or their combinations, without any tailored instruction-following tuning.
  }
  \label{fig:teaser}
\end{figure}

Omni-modal representation learning has emerged as a focal point of research, garnering a surge of interest owing to its relevance in real-world applications like Embodied AI.
In the uni-modal realm, the community has harnessed the versatility of Transformers~\cite{vaswani2017attention} in conjunction with the abundance of large-scale web data, yielding notable successes in the development of general-purpose foundation models for Natural Language Processing~\cite{bert,roberta,gpt1,gpt2,gpt3,openai2022chatgpt} and Computer Vision~\cite{dosovitskiy2020image,he2022masked,bao2021beit,peng2022beitv2,fang2023eva,fang2023eva02}.
% , such as Language Models~\cite{bert,roberta,gpt1,gpt2,gpt3,openai2022chatgpt} in Natural Language Processing (NLP), and Vision Foundation Models~\cite{dosovitskiy2020image,he2022masked,bao2021beit,peng2022beitv2,fang2023eva,fang2023eva02} in Computer Vision (CV).
In cross-modal research, the vision-language model CLIP~\cite{openai_clip} stands as a significant milestone by capitalizing on large-scale image-text paired data for pertaining.
% , CLIP has successfully bridged the semantic gap between language and vision.
% By capitalizing on large-scale image-text paired data for pretraining, CLIP has successfully bridged the gap between language and vision learning, propelling the SOTA in computer vision research with the highly effective vision model.
While proficient in integrating vision and language, the present CLIP-based models are constrained in their ability to encompass and interpret additional sensory modalities, such as audio, depth, and 3D shape.

% However, despite these remarkable advancements, the endeavor to achieve comprehensive omni-modal representation learning remains a challenge. 
% Despite these remarkable advancements, achieving comprehensive omni-modal representation learning remains challenging. 
% The present foundation models, while proficient in integrating vision and language, are constrained in their ability to encompass and interpret additional sensory modalities, such as audio, depth, and 3D shape. 
Several recent works, \textit{e.g.}, ImageBind~\cite{girdhar2023imagebind} and ONE-PEACE~\cite{wang2023onepeace}, have initiated the pursuit of constructing a representation model that can encompass a wide range of modalities. 
With either modality-specific encoders or a modality-shared architecture, they heavily rely on large-scale data with cross-modal aligned semantics (\textit{e.g.}, paired with image or text), which inevitably hinders their ability for accommodating an extensive variety of modalities.
% For example, ImageBind~\cite{girdhar2023imagebind} proposes to learn a shared representation space by leveraging multiple types of image-paired data, where separate encoders are trained for each modality to align the image representation~\cite{openai_clip,cherti2022openclip}.
% of a pretrained Vision-Language model~\cite{openai_clip,cherti2022openclip}. 
% ONE-PEACE~\cite{wang2023onepeace} introduces an architecture with modality adapters and a fusion encoder, training the model from scratch using large-scale image-text and audio-text paired data to integrate representations across vision, audio, and language.
% Despite the progress made by these works, achieving true omni-modal representations still poses challenges. The existing approaches are not entirely scalable for accommodating an extensive variety of modalities due to their heavy reliance on large-scale data for pretraining~\cite{wang2023onepeace}. 
% separate encoders for different modalities~\cite{girdhar2023imagebind}, or 
% the necessity of large-scale data for pretraining~\cite{wang2023onepeace}. 
Specifically, gathering large-scale data for certain modalities can be non-trivial. This leads to sub-optimal models with poor generalization when facing novel categories, thus hindering their wider applications in the real world.

% Nevertheless, given the exceptional generalization and transfer learning capabilities of the \ptvit models~\cite{openai_clip,fang2023eva,fang2023eva02,caron2021dino,oquab2023dinov2}, there is promise in adapting the inherent knowledge to comprehend novel modalities, without the need for collecting large-scale data and training distinct models from scratch for each modality, which demands substantial time and resources.
% \textcolor{red}{without the requirement of xxx and xxx}. 
% To obtain knolwedge from the pretrained models, prior works mainly follow the pretrain-and-then-finetune or knowledge distillation paradigms~\cite{girdhar2023imagebind,luo2022clip4clip,fang2023eva,xue2023ulip,xue2023ulip2,liu2023openshape}.

In this work, we present a novel perspective. 
Given the exceptional generalization and transfer learning capabilities of the \ptvit models~\cite{openai_clip,fang2023eva,fang2023eva02,caron2021dino,oquab2023dinov2}, there is promise in adapting the inherent knowledge to comprehend novel modalities, without the need for collecting large-scale data to train models from scratch for each modality, which demands substantial time and resources. Recognizing the rich knowledge encoded in a \ptvit, we conjecture that a \ptvit is able to function as a multi-modal processor -- it possesses the capacity to sense and comprehend a spectrum of modalities as it interprets images. 
% Insights from cognitive science underscore that the human brain adeptly processes and integrates diverse modalities, encompassing vision, hearing, touch, and more, within dedicated brain regions~\cite{calvert2001crossmodal,calvert2004multisensory}.
% Drawing inspiration from this phenomenon, and recognizing the comprehensive knowledge encompassed within a \ptvit, we conjecture that a \ptvit similarly operates as a multi-modal sensor -- it possesses the capacity to sense and comprehend a spectrum of modalities as it perceives and understands images. 

From this standpoint, 
% rather than distilling knowledge by aligning the high-level feature space (\emph{i.e.}, the feature of the \texttt{[CLS]} token or mean/max pooling feature from the last layer of the transformer) of a vision/language foundation model and a modality-specific encoder, 
we introduce \methodname, which encodes the out-of-image modalities through a \ptvit, with the goal of maximizing the utilization of pretrained model weights and the knowledge they encapsulate.
As illustrated in Figure~\ref{fig:teaser}, \methodname\ integrates a modality embedding module and a perceiver~\cite{jaegle2021perceiver} to map input data into the \ptvit input space. Subsequently, a frozen \ptvit is applied for further encoding. This approach enables the encoding of diverse modalities, aligning their features with the established features of anchor data, be it images, text, or image-text, through an off-the-shelf foundation model.
% The output embedding is then learned to align the off-the-shelf CLIP features of paired images, text, or image-text through multi-modal contrastive learning.

% what is the advantage
Our proposed method offers several advantages in advancing omni-modal representation learning:
(1) \textbf{Model Unification.} Our \methodname adopts a shared \ptvit for various modalities, facilitating scalable modality extension and aligning with the growing trend of big convergence in multi-modal understanding~\cite{wang2023beit3}.
(2) \textbf{Data-Efficient Approach}. \methodname achieves its versatility and applicability across diverse tasks and domains by effectively utilizing the advanced \vit model without the need for large-scale data.
(3) \textbf{Emergent Ability.} Recent studies in Multimodal Large Language Models (MLLMs)~\cite{blip2,alayrac2022flamingo,openai2023gpt4,zhu2022minigpt4,liu2023llava,dai2023instructblip} have underscored the essential role of vision representation models to unleash Large Language Models (LLMs) to perceive and understand the visual world. 
Interestingly, we find that by binding a new modality to the \vit used in an off-the-shelf MLLM, the corresponding LLM can seamlessly sense and comprehend that modality without any specific instruction tuning.
% This observation highlights the adaptability and innate capabilities of \methodname in effortlessly integrating and comprehending novel modalities.

To demonstrate the effectiveness of \methodname, we evaluate its performance in the context of 3D shape understanding.
Specifically, we follow the pre-training paradigms of prior works~\cite{xue2023ulip,xue2023ulip2,liu2023openshape}, but aims to establish a more generalized and scalable joint representation space encompassing 2D images, text and 3D shapes by binding 3D point cloud to a \ptvit. 
On the zero-shot 3D shape classification task, \methodname exhibits substantial improvements over previous zero-shot state-of-the-art methods.  Notably, using the same datasets for pretraining, \methodname outperforms ULIP~\cite{xue2023ulip} by \textbf{10.2\%}, ULIP2~\cite{xue2023ulip2} by \textbf{10.4\%}, and OpenShape~\cite{liu2023openshape} by \textbf{3.2\%} on ModelNet40~\cite{wu2015modelnet} in terms of zero-shot accuracy.
\methodname also excels at handling long-tail categories. On the challenging Objaverse-LVIS dataset~\cite{deitke2023objaverse} containing 1,156 categories, \methodname achieves a \textbf{52.0\%} zero-shot accuracy, significantly surpassing previous SOTA by \textbf{5.2\%}.

Besides zero-shot classification, we bind 3D shapes to the \vit architecture employed in InstructBlip~\cite{dai2023instructblip}, an MLLM capable of understanding and interacting over 2D images. After pretraining, we plug the 3D encoder produced by \methodname into InstructBlip to investigate whether \methodname can bestow the LLM with the ability to perceive and comprehend the 3D modality.
% plug the 3D encoder produced by \methodname into InstructBlip~\cite{dai2023instructblip}, an MLLM capable of understanding and interacting over 2D images, \textcolor{red}{to ....}. 
This integration showcases that tthe new variant of the MLLM is endowed with the capability for 3D shape captioning and question answering, without necessitating specific instruction tuning.

\methodname aims to pursue omni-modal representation learning in a simple yet effective manner, alleviating the need for large-scale data collection and utilizing a single set of knowledge expert parameters.
Our initial exploration strives to enable Large Language Models (LLMs) to sense and comprehend out-of-image modalities in a zero-shot manner. Future work can further scale up training to extend \methodname to more modalities and explore additional emergent abilities.

\section{\methodname for 3D Shape Understanding}
\methodname assimilates the pre-training framework in~\cite{xue2023ulip,xue2023ulip2,liu2023openshape} and introduces to learn 3D shape representation through a \ptvit. In this context, we use \clipvit, which allows \methodname to leverage the rich knowledge from the large-scale image-text data, adopting the transferalbility of the original \clipvit model. 
By extending this paradigm to encompass more modalities, \methodname not only enables a \vit to seamlessly sense and comprehend 3D shapes but also allows it to understand other modalities as it digest images, all while utilizing a single set of model parameters.

\subsection{Preliminary: CLIP for 3D Shape Representation Learning}
CLIP (Contrastive Language-Image Pretraining) is a powerful multimodal model trained on a large dataset of image-text pairs. It learns to map images and their corresponding text into a shared embedding space through contrastive learning. 
The resulting joint image-text embeddings enable CLIP to perform various multimodal tasks, showcasing remarkable generalization and zero-shot learning capabilities. 
CLIP's knowledge, acquired from extensive pretraining on image-text data, encapsulates rich information about the visual world and language. This inherent knowledge makes CLIP highly transferable, allowing it to excel in downstream tasks and domains without task-specific fine-tuning. 

Building upon CLIP, ULIP~\cite{xue2023ulip} introduces an efficient multimodal pretraining framework using triplets that encompass three modalities: (1) the 3D modality, extracted from 3D point cloud data; (2) the image modality, generated by rendering images from 3D object files with multiple viewpoints; and (3) the language modality, derived by converting dataset metadata into coherent sentences, including descriptive terms and category names.  
Subsequently, ~\cite{xue2023ulip2,liu2023openshape} scale up the pretraining 3D data and leverage a large language model~\cite{blip,blip2,openai2023gpt4} to automatically generate detailed captions from a comprehensive set of  holistic views, reducing reliance on human annotations. All these methods learn 3D shape representations aligned with the pretrained CLIP embedding spaces of language and image.
\methodname follows the same training paradigm as these works but incorporates the pretrained vision part for 3D shape encoding, aiming to utilize the \clipvit from the model perspective.

\begin{figure}[th]
  \centering
  \includegraphics[width=0.80\textwidth]{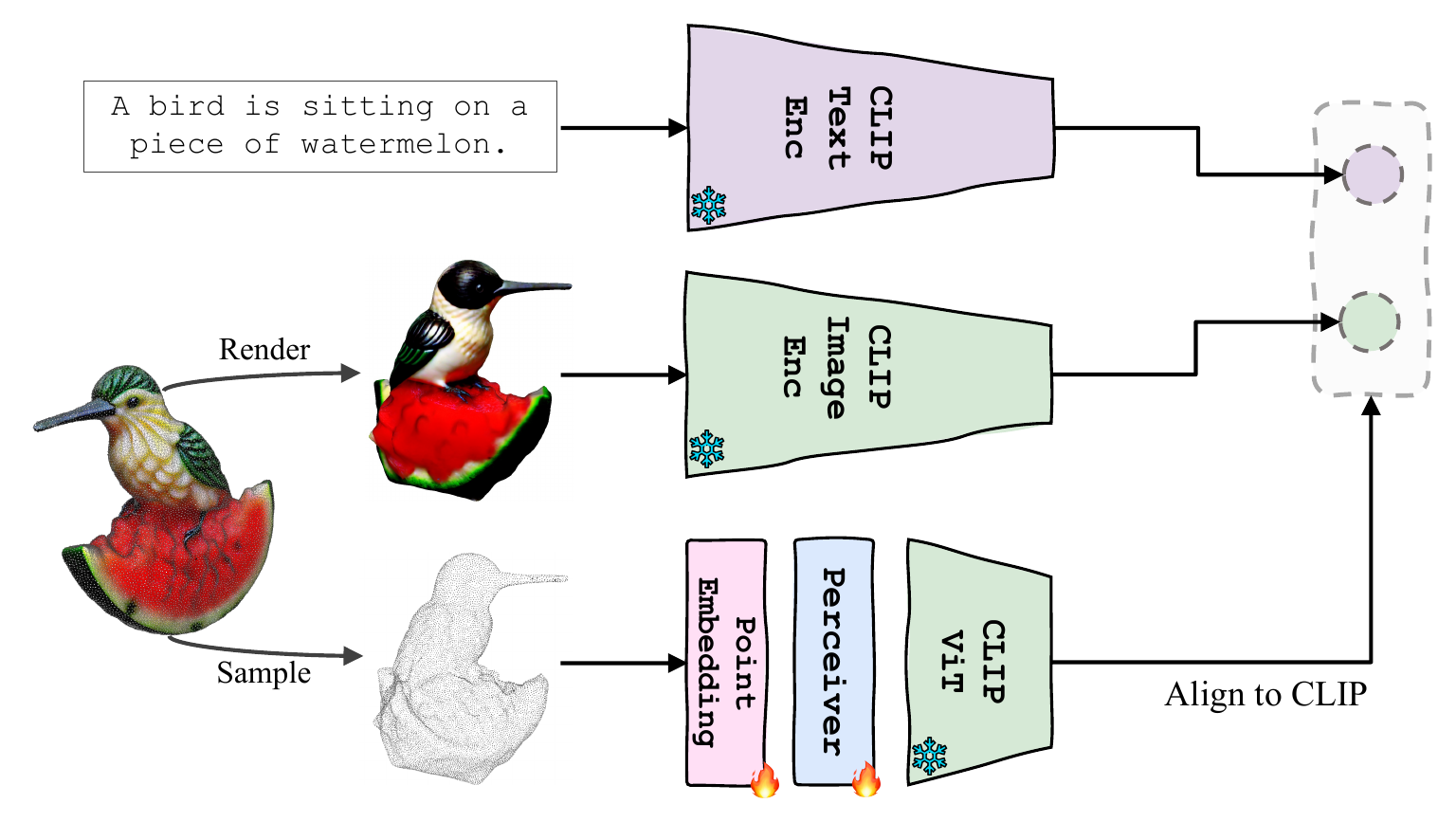}
  \caption{\textbf{Training pipeline of \methodname for 3D shape understanding.} \methodname aligns the triplet of 3D point clouds, 2D rendered images, and textual descriptions to a unified feature space, defined by CLIP.
  % \textcolor{red}{defined by CLIP}. 
  It leverages the capabilities of a powerful pretrained vision language model, CLIP~\cite{openai_clip,cherti2022openclip}, which is frozen during pretraining and provides a pre-aligned feature space. 
  The 3D shape encoder consists of a point embedding layer, a Perceiver, and a pretrained \clipvit, shared with the image encoder. To enhance 3D point cloud encoding, point embeddings are obtained and distilled through the Perceiver before feeding into the frozen \clipvit to obtain the final representation.
  % Notably, the Perceiver enables mapping input signals from various modalities into the input space of \clipvit, allowing \clipvit to sense and comprehend 3D data. 
  The training objective is to minimize the contrastive loss for aligning features in the shared feature space.}
  \label{fig:3d-arch}
\end{figure}

\subsection{Architecture}
\textbf{Text and Image Encoders.} \methodname aligns the triplet of 3D point clouds, 2D rendered images, and comprehensive descriptions to a unified feature space. As depicted in Figure~\ref{fig:3d-arch}, we leverage the capabilities of a powerful pretrained vision language model, CLIP~\cite{openai_clip,cherti2022openclip}, and freeze it during pretraining. The feature space, which has already been pre-aligned by CLIP, serves as the target space where we aim to  integrate the 3D modality.

\textbf{3D Shape Encoder.} As is shown in Figure~\ref{fig:3d-arch}, the model architecture of 3D encoder in \methodname consists of a point embedding layer~\cite{yu2022pointbert}, a Perceiver~\cite{jaegle2021perceiver} and a pretrained \clipvit. 
Due to the distinct characteristics of different modalities, directly inputting the 3D point cloud into the pretrained \clipvit can lead to a mismatch in the input space, resulting in suboptimal performance even with the use of a powerful model.
Therefore, we employ some heuristic designs prior to \clipvit for better 3d point cloud encoding: (1) \textit{Obtain point embeddings:} we first partition the input point clouds into several point patches (sub-clouds) and then map them into a sequence of point embeddings~\cite{yu2022pointbert}; (2) \textit{Map point embeddings into \clipvit input space:} we apply the Perceiver architecture~\cite{jaegle2021perceiver} to iteratively distill the point embeddings into a group of latent embeddings, thereby constructing the input for \clipvit.
% \textcolor{red}{to ...}.
Subsequently, the latent embeddings are forwarded to the frozen \clipvit to obtain the final 3D representation.

During pretraining, \clipvit remains frozen, and only the parameters of point embedding and Perceiver are updated. This design philosophy is equally applicable across other modalities.  This modulated architecture not only retains the foundational knowledge ingrained within the \vit model, but also realizes computational efficiency in comparison to training the entire encoder.

\textbf{Perceiver: Connecting Modalities to \clipvit.} Notably, the Perceiver architecture leverages an asymmetric attention mechanism to iteratively distill inputs into a tight latent bottleneck, allowing it to handle very large inputs of arbitrary sizes and thus can accommodate different modalities.  Similar architectures are employed in Vision-Language Models (VLMs) like Flamingo~\cite{alayrac2022flamingo} and BLIP-2~\cite{blip2} to extract visual information for Large Language Models (LLMs). 
However, \methodname takes a novel approach by using the Perceiver to map input signals from various modalities into the input space of the pretrained \clipvit, unleashing \clipvit to sense and comprehend other modalities beyond images. This innovative design enables \methodname to achieve omni-modal representations in a straightforward and efficient manner.

\textbf{Multimodal Representation Alignment}.
We train the 3D encoder that takes  3D point clouds as input and extracts 3D shape feature. Following the approach in previous works~\cite{xue2023ulip,xue2023ulip2,liu2023openshape}, we adopt multimodal contrastive learning for representation alignment. Given the 3D encoder $F^P$, the frozen CLIP image encoder $F^I$, and the frozen CLIP text encoder $F^T$, along with a sampled batch of triplets $\left\{ (P_i, I_i, T_i)\right\}$ for the 3D point clouds of a shape, its corresponding rendered image, and the associated text, the contrastive loss for alignment is formulated as:
\begin{multline*}
    \mathcal{L} = -\frac{1}{4B}\sum_{i=1}^B \left(
    \underbrace{
    \log \frac{\exp(h_i^P \cdot h_i^I / \tau)}{\sum_j \exp (h_i^P \cdot h_j^I / \tau)} 
    + \log \frac{\exp(h_i^I \cdot h_i^P / \tau)}{\sum_j \exp (h_i^I \cdot h_j^P / \tau)}}_{\mathcal{L}_{P2I}\text{: Point Cloud-Image contrastive}} \right. \\ 
    \left.
    + \underbrace{\log \frac{\exp(h_i^P \cdot h_i^T / \tau)}{\sum_j \exp (h_i^P \cdot h_j^T / \tau)} 
    + \log \frac{\exp(h_i^T \cdot h_i^P / \tau)}{\sum_j \exp (h_i^T \cdot h_j^P / \tau)}}_{\mathcal{L}_{P2T}\text{: Point Cloud-Text contrastive}}
    \right),
\end{multline*}
where $B$ is the number of shapes in a batch; $\tau$ is a learnable temperature; $h_i^P = {F^P(P_i)} / {\|F^P(P_i)\|}$, $h_i^I = {F^I(I_i)}/{\|F^I(I_i)\|}$ and $h_i^T = {F^T(T_i)} / {\|F^T(T_i)\|}$ denote normalized features of $P_i$, $I_i$ and $T_i$.
The objective of training the 3D encoder is to minimize $\mathcal{L}$.

% ---------------------------------------------------------------
\section {Experiments}
\subsection{Experimental Setting}\label{sec:exp_setting}
\textbf{Pretraining Datasets.} Our experimental setup for pretraining leverages datasets from prior works~\cite{xue2023ulip,xue2023ulip2,liu2023openshape} at varying scales:
\dsA \textbf{ULIP-ShapeNet Triplets}~\cite{xue2023ulip} are derived from ShapeNet55~\cite{chang2015shapenet}, where point clouds are generated from CAD models. Images are synthesized using virtual cameras positioned around each object, and texts are obtained by filling metadata into a predefined prompt template.
\dsB \textbf{ULIP2-Objaverse Triplets}~\cite{xue2023ulip2} utilize the recently released Objaverse~\cite{deitke2023objaverse}. For each 3D object, 12 rendered images are used, spaced equally by 360/12 degrees. Each rendered image has 10 detailed captions generated using BLIP2-opt6.7B~\cite{blip2}.
\dsC \textbf{OpenShape Triplets}~\cite{liu2023openshape} encompass four prominent public 3D datasets: ShapeNet~\cite{chang2015shapenet}, 3D-FUTURE~\cite{fu2021-3d-future}, ABO~\cite{collins2022abo}, and Objaverse~\cite{deitke2023objaverse}. For each 3D object, 12 color images are rendered from preset camera poses, and thumbnail images are included as candidates if provided. OpenShape employs various strategies to obtain high-quality text descriptions, including filtering noisy metadata using GPT4~\cite{openai2023gpt4}, generating captions using BLIP~\cite{blip} and Azure cognition services, and conducting image retrieval on LAION-5B to retrieve relevant texts with paired images closely resembling the object's rendered image, leading to a wider range of text styles.
We show the statistics of the pretraining datasets in Table.\ref{tab:pt_dataset}.

\begin{table}[h]
\begin{minipage}{.53\linewidth}
\centering

\caption{Statistics of pretraining datasets.}\label{tab:pt_dataset}
\resizebox{\textwidth}{!}{%
\begin{tabular}{c|c|c}
\hline
Dataset & Source                                                                       & \# 3D point clouds \\ \hline \hline
\begin{tabular}[c]{@{}c@{}} \dsA \\ ULIP-ShapeNet Triplets\end{tabular}      & ShapeNet~\cite{chang2015shapenet}                                                                     & $\sim$52.5k  \\ \hline
\begin{tabular}[c]{@{}c@{}} \dsB \\ ULIP2-Objaverse Triplets\end{tabular}     & Objaverse~\cite{deitke2023objaverse}                                                                    & $\sim$798.8k \\ \hline
\begin{tabular}[c]{@{}c@{}} \dsC \\ OpenShape Triplets\end{tabular}     & \begin{tabular}[c]{@{}c@{}}ShapeNet~\cite{chang2015shapenet}, \\ 3D-FUTURE~\cite{fu2021-3d-future},\\ ABO~\cite{collins2022abo}, \\ Objaverse~\cite{deitke2023objaverse}\end{tabular} & $\sim$876k  \\ \hline
\end{tabular}
}

\end{minipage}%
\hspace{.02\linewidth}
\begin{minipage}{.43\linewidth}
\centering
\caption{CLIP models used in experiments. } \label{tab:clip-models}
\vspace{0.3cm}
\resizebox{\textwidth}{!}{%
\begin{tabular}{l|l|l}
\hline
\multicolumn{1}{c|}{CLIP Model} & \multicolumn{1}{c|}{Source} & \multicolumn{1}{c}{CLIP PT Dataset} \\ \hline \hline
OpenAI-B16                      & OpenAI                      & WIT-400M                        \\
Laion2B-B16                     & OpenCLIP                    & LAION-2B                        \\
OpenAI-L14                      & OpenAI                      & WIT-400M                        \\
Datacomp-L14                    & OpenCLIP                    & DataComp-1B                     \\
EVA01-g14                       & BAAI                        & LAION-400M                      \\
bigG14                          & OpenCLIP                    & LAION-2B  \\ \hline                     
\end{tabular}
}
\end{minipage}
\vspace{-0.1cm}
\end{table}

\textbf{Downstream Datasets.} We use the following datasets for downstream tasks.
(i) ModelNet40~\cite{wu2015modelnet} is a synthetic dataset of 3D CAD models with 9,843 training samples and 2,468 testing samples, covering 40 categories.
(ii) ScanObjectNN~\cite{uy2019scanobjectnn} is a dataset of scanned 3D objects from the real world, containing 2,902 objects categorized into 15 categories. We follow~\cite{xue2023ulip,xue2023ulip2,liu2023openshape} and use the variants provided by~\cite{yu2022pointbert} in our experiments.
(iii) Objaverse-LVIS is an annotated subset of Objaverse~\cite{deitke2023objaverse}, comprising 46,832 shapes from 1,156 LVIS~\cite{gupta2019lvis} categories. With a larger base of classes compared to other benchmarks, Objaverse-LVIS presents a challenging long-tailed distribution, making it a better reflection of the model's performance in open-world scenarios.

\textbf{Implementation Details.}
In our 3D shape understanding experiments, we employ CLIP models~\cite{openai_clip,cherti2022openclip} to encode text descriptions and rendered images. The specific CLIP models used in our experiments are outlined in Table~\ref{tab:clip-models}. The \clipvit version, serving as the CLIP image encoder, is directly integrated into the 3D encoder \textit{\textbf{by default}}.
% For the Perceiver component, we utilize 4 Perceiver blocks, each comprising 1 cross-attention module and 6 self-attention modules. These Perceiver blocks share parameters across the entire architecture. The dimensions of Perceiver latent arrays and the number of attention heads are set according to the adopted \clipvit version.

Regarding data preprocessing, we uniformly sample 8,192 points for \dsA ULIP-ShapeNet Triplets, \dsB ULIP2-Objaverse Triplets, and 10,000 points for \dsC OpenShape-Triplets. These 3D inputs are further partitioned into 512 point patches (sub-clouds), where each sub-cloud contains 32 points. To achieve this, we first perform the Farthest Point Sampling (FPS) operation to sample a representative skeleton and then employ the K-Nearest Neighbor (KNN) method to group neighboring points.
For text descriptions and rendered images, we follow the pre-processing and augmentation procedures in CLIP~\cite{openai_clip,cherti2022openclip}. 

During pretraining, we freeze the text encoder, image encoder, and \clipvit in the 3D encoder, updating only the parameters of modality-specific embeddings and the Perceiver.
% For optimization, we use the AdamW optimizer with a global batch size of 512.

\subsection{Zero-Shot 3D Classification}
\subsubsection{Comparison with State-of-the-arts}
\input{tab/tab_zs_3dcls}
We conducted zero-shot classification evaluations of our models on three datasets: ModelNet40~\cite{wu2015modelnet}, ScanObjectNN~\cite{uy2019scanobjectnn}, and Objaverse-LVIS~\cite{deitke2023objaverse}. To ensure a fair comparison, we used the same datasets for pretraining as in previous works~\cite{xue2023ulip,xue2023ulip2,liu2023openshape}. The comprehensive results can be found in Table~\ref{tab:zero-shot-all}.

In particular, we compared \methodname with ULIP~\cite{xue2023ulip}, using PointNet++~\cite{qi2017pointnet++}, PointMLP~\cite{ma2022pointmlp}, and PointBERT~\cite{yu2022pointbert} as 3D encoders. Table~\ref{tab:zero-shot-ulip} presents the outcomes when our model is pretrained on \dsA ULIP-Triplets. Notably, various variants of CLIP models employed in \methodname outperform ULIP with different backbones. Particularly, using \clipvit-L14 (Datacomp)~\cite{cherti2022openclip} achieves a top-1 accuracy of 70.7\% and a top-5 accuracy of 92.8\% on ModelNet40, even surpassing ULIP2-PointBERT pretrained on the larger \dsB ULIP2-Objaverse-Triplets.

The results of pretraining on \dsB ULIP2-Triplets are presented in Table~\ref{tab:zero-shot-ulip2}. \methodname with different variants significantly improves the zero-shot classification accuracy compared to ULIP2-PointNeXt~\cite{qian2022pointnext} and ULIP2-PointBERT~\cite{yu2022pointbert}. Notably, \methodname with \clipvit-L14 (Datacomp)~\cite{cherti2022openclip} outperforms ULIP2-PointBERT by 10.4\% in top-1 accuracy.

In Table~\ref{tab:zero-shot-ops}, we present the results of pretraining on the large-scale and text-enriched \dsC OpenShape-Triplets. To align with OpenShape~\cite{liu2023openshape} and the released data, we adopted \clipvit-bigG14 from Open CLIP~\cite{cherti2022openclip} for \methodname and trained on both \dsC NO LVIS (excluding all shapes from the Objaverse-LVIS subset) and \dsC All for comparison. \methodname outperforms models adopted in OpenShape, even when their backbones are scaled for improved performance. 
Notably, \methodname significantly improves the classification accuracy on the long-tail categories of Objaverse-LVIS, from 46.8\% to 52.0\%. Additionally, when pretrained on the NO LVIS subset, \methodname achieves a top-1 accuracy of 50.2\%. This outperformance is evident as it surpasses ULIP by approximately 30\%, and even surpasses OpenShape-PointBERT, trained on the entire set, by 3.3\%.  These results demonstrate \methodname's capability to recognize open-world objects effectively.
Regarding ModelNet40, \methodname achieves an 87.4\% accuracy, surpassing previous SOTAs on zero-shot classification and outperforming the fully-supervised 3D ShapeNets~\cite{wu2015modelnet} and VoxNet~\cite{maturana2015voxnet}. Additionally, on ScanObjectNN, containing challenging real scans with noise and occlusion, our method exhibits decent sim-to-real transfer capabilities. \methodname achieves a 60.6\% zero-shot accuracy on ScanObjectNN without specific sim-to-real training, surpassing the previous SOTA.

\subsubsection{Ablation Study}
We perform various ablations by pretraining on \dsA ULIP-ShapeNet Triplets~\cite{xue2023ulip} and evaluating on the ModelNet40~\cite{wu2015modelnet} zero-shot classification benchmarks, unless otherwise specified. The comprehensive results are presented in Table~\ref{tab:abla-all} and the default setting is marked with \colorbox{lavenderweb}{color}, if applicable.

\textbf{Comparison with PointBERT.} We conduct experiments to compare \methodname with PointBERT~\cite{yu2022pointbert}, a transformer based architecture for 3D point cloud understanding. This comparison involves aligning to the feature space of different CLIP variants and employing distinct pretraining datasets (refer to Table~\ref{tab:pt_dataset} and Table~\ref{tab:clip-models}).
As is shown in Table~\ref{tab:abla-3d-pointBERT}, \methodname outperforms PointBERT over all combinations of pretraining datasets and CLIP model for alignment. \methodname consistently outperforms PointBERT across all conceivable combinations of pretraining datasets and CLIP models for alignment. This substantiates the efficacy of harnessing a pretrained \vit to advance 3D shape understanding.

\input{fig/graphs/model_dataset_openai}
\textbf{Scaling pretraining dataset and model size.}
We investigate the effects of model size and pretraining data scale. Specifically, we choose OpenAI-B16 and OpenAI-L14~\cite{openai_clip}, both pretrained on WIT-400M, to ablate the model size. We conduct training on \dsA ULIP-ShapeNet Triplets and \dsB ULIP2-Objaverse Triplets to analyze the effect of dataset scale.  The results in Figure~\ref{fig:abla-scaling1} reveal that while keeping the dataset size constant, elevating the model scale from B16 to L14 enhances zero-shot performance, particularly noticeable for the larger L14 model. Moreover, when the model size is fixed, expanding the pretraining dataset yields improvements in model performance.

\input{fig/graphs/scale_cmp_pointbert}
\textbf{Comparing Scaling Capabilities of PointBERT and \methodname.} We evaluate the scalability of PointBERT~\cite{yu2022pointbert} and \methodname. Our experiments, depicted in Figure~\ref{fig:abla-scaling2}, involve varying both model sizes and pretraining datasets along the x-axis, ranging from \dsA::Laion2B-B16 and \dsB::Datacomp-L14 to \dsC::bigG14. Here, the data scaling encompasses increases in dataset size and data quality.  When scaling the model size of \methodname, we correspondingly scale up the PointBERT model. The results shown in Figure~\ref{fig:abla-scaling2} underscore that \methodname consistently outperforms PointBERT across all scaling configurations, owing to the rich knowledge encoded in the \ptvit. 
Notably, when PointBERT is scaled to a large size (with 171M parameters), it achieves a top-1 zero-shot accuracy of 80.2\%, lower than the use of a smaller PointBERT (with 32M parameters) for representation learning (\emph{i.e.}, 84.4\% zero-shot accuracy), similar to the findings in~\cite{liu2023openshape}. This substantiates that \methodname excels in scalability.

\textbf{Configuration of Perceiver in \methodname.} We study the effect of different design choices concerning the Perceiver~\cite{jaegle2021perceiver} employed in \methodname. Our study encompasses the ablation of Perceiver depth, representing the number of Perceiver blocks, as well as the exploration of parameter sharing beyond the second block (included).
Results in Table~\ref{tab:abla-perceiver} show that increasing the depth of Peiceiver does not yield enhanced performance. Additionally, sharing parameters among Perceiver blocks proves capable of reducing parameters while achieving comparable or superior performance. This underscores the effectiveness and efficiency of the Perceiver architecture within \methodname for connecting the 3D input to a \ptvit.

\input{tab/tab_abla}

\textbf{Other hyper-parameters in \methodname.} We vary the number of latents used in the Perceiver architecture, which aligns with the sequence length of the \ptvit input. As illustrated in Table~\ref{tab:abla-hyperparam}, employing a larger number of latents, such as 384 and 512, leads to slightly improved performance while concurrently increasing the GFlops.  
In this regard, we are able to harness the intrinsic capability of the Perceiver to extract insights from inputs of variable sizes and seamlessly connect them to the \ptvit, thereby mitigating computational complexity.
Furthermore, we investigate whether the inclusion of the \ptvit position embedding influences model performance. In particular, we examine the case of CLIP Laion2B-B16 by interpolating the original position embedding while varying the number of latents. The outcomes presented in Table~\ref{tab:abla-hyperparam} indicate that omitting the pretrained position embedding does not significantly degrade performance. We conjecture that the Perceiver has the capacity to implicitly assimilate position information on its own.

\textbf{Point Embedding $\rightarrow$ Perceiver.} To validate the efficacy of the \ptvit, we investigate the performance of the ``Point Embedding $\rightarrow$ Perceiver'' paradigm. In this setup, the mean pooling feature of the Perceiver is directly aligned with the CLIP feature space. We execute experiments with various hyperparameter configurations, and the comprehensive outcomes are presented in Table~\ref{tab:abla-point-perceiver}.
Typically, the configuration employing a depth of 6 without parameter sharing possesses a comparable total parameter count to the default setting of \methodname ($\sim$119M). Despite having less trainable parameters, \methodname outperforms this variant by a significant margin. This observation underscores the pivotal role of harnessing the capabilities of the \ptvit.

\textbf{Point Embedding $\rightarrow$ \ptvit.} We also delve into the paradigm of ``Point Embedding $\rightarrow$ \ptvit''. As depicted in Table~\ref{tab:abla-point-vit}, only training the Point Embedding yields a zero-shot accuracy of 50\%, significantly lower than that of \methodname due to the restricted number of trainable parameters. 
Furthermore, unlocking the transformer blocks for training leads to improved zero-shot performance. However, this tailored training approach optimized for 3D understanding might limit the adaptability of the resultant \vit to other modalities, thereby diminishing the \ptvit's overall generalization ability. 
Contrastingly, \methodname attains commendable performance while largely retaining the core parameters of the \ptvit, effectively harnessing its expansive knowledge across a wide array of modalities, with only a marginal increase in new parameters.

\subsection{Emergent Capability: Connecting \methodname to LLM Without Training}
In this section, we study whether \methodname's representation can be used to unleash an LLM to understand the out-of-image modalities. Specifically, in this work, we use the 3D data for pilot experiment.

\textbf{Prelinminary for MLLM}. Recent advancements in large language models (LLMs) have demonstrated remarkable emergent abilities, yet they lack the capacity to comprehend other modalities beyond text. Recent representation models have shown their bedrock role to unleash LLMs to understand, perceive, and interact with other modalities. For instance, Multimodal Large Language Models (MLLMs) effectively combine the LLMs and vision models, enabling reasoning with both textual and visual data. MLLMs not only provide human-like perception but also facilitate user-friendly interactions and a broader scope for solving diverse tasks.

\input{tab/tab_mllm_example_plant}
\input{tab/tab_mllm_example_piano}
\input{tab/tab_mllm_example_toilet}

\textbf{Experimental Details of Unimodal Inputs.} We use InstructBLIP~\cite{dai2023instructblip}, which is a versatile framework that enhances general-purpose models for a wide range of visual language tasks using a unified natural language interface by employing diverse instruction data to train an MLLM. InstructBLIP applies the pretrained EVA01-g14~\cite{fang2023eva} \clipvit to perceive and encode images. 
Adhering to InstructBLIP's configuration, we employ the EVA01-g14 CLIP for both images and texts, utilizing its \clipvit as an integral part of the 3D encoder for point cloud encoding. We fine-tune the parameters of the point embedding and Perceiver on \dsB ULIP2-Objaverse Triplets. Following training, we seamlessly integrate \methodname into InstructBLIP, enabling the resulting MLLM to effectively handle 3D input.

% \textbf{Qualitative Results on Unimodal Inputs.} 
Our comparative analysis encompasses: (1) PointBERT aligned with the same CLIP model, replacing the post-training vision encoder in InstructBLIP; and (2) CLIPCap~\cite{mokady2021clipcap} from OpenShape~\cite{liu2023openshape}.
We offer a snapshot of qualitative outcomes across different models in Table~\ref{tab:visual_example_plant}, Table~\ref{tab:visual_example_piano}, and Table~\ref{tab:visual_example_toilet}. These examples spotlight several capabilities exhibited without specific MLLM tuning using 3D-related instruction data. Notably, the examples demonstrate that \methodname empowers InstructBLIP to accurately describe 3D objects. For instance, the plant example in Table~\ref{tab:visual_example_plant} is characterized as ``sitting in a ceramic pot'' and ``bamboo-like''. Moreover, \methodname excels in capturing nuanced visual concepts beyond the most prominent ones. For instance, the piano example reveals the observation of a ``chair''.

For InstructBLIP w/ PointBERT, although PointBERT is able to conduct zero-shot classification, it fails to provide accurate information for the LLM to digest in such a zero-shot manner, and thus, InstructBLIP w/ PointBERT can not describe the 3D object correctly. 
CLIPCap-OpenShape, while displaying some relevant entities in captions (``vase'' in Table~\ref{tab:visual_example_plant} and ``toilet'' in Table~\ref{tab:visual_example_toilet}), often leads to hallucinations and inaccurate captions.

The overall results demonstrate that \methodname excels not only at classifying the salient object of the 3D input, but also capturing the visual details, potentially attributed to the robustness of the \ptvit,  bringing the emergent ability to integrate with off-the-shelf LLM in a zero-shot manner.

% mm3: line 434,386,338,506,578,987,1035,1491,1683,1778
% mm2 ,939,
\textbf{Experimental Details for Multimodal Inputs.} Following the previous setting, we plug the \methodname into the InstructBLIP. Here, we show that a \ptvit equipped with a variety of ``lens'' is a multimodal sensor that can perceive and integrate multimodalities simultaneously. To achieve this, we concatenate outputs from diverse modal lenses before the \vit transformer. The \ptvit then encodes these concatenated embeddings and transmits the result to InstructBLIP for conditioned text generation. Our specific experiment involves the fusion of image~\footnote{Photos credited to \url{https://www.pexels.com/}.} and point cloud data. 

Qualitative outcomes are presented in Table~\ref{tab:example_mm2_insblip} for dual multimodal inputs (a point cloud instance and an image) and in Table~\ref{tab:example_mm3_insblip} for tri-modal inputs (a point cloud instance and 2 images). The responses from InstructBLIP w/ \methodname underscore its adaptability in seamlessly integrating diverse modalities, allowing InstructBLIP to interpret multimodal inputs as it is viewing an image. Notably, this integration empowers the resulting MLLM to depict multimodal inputs, discern unconventional co-occurrence of concepts from different modalities, and craft narratives based on the integrated information of multimodalities.

\begin{table*}[h!]\centering
\begin{minipage}{1.0\columnwidth}\vspace{0mm}
\centering
\caption{Example to illustrate the Instruct-BLIP with two multimodal inputs.}
    \label{tab:example_mm2_insblip}
\begin{tcolorbox}[colback=white,colframe=blue!75!black,title={\bf InstructBLIP w/ \methodname, Two Multimodal Inputs.}]
    \centering
      \footnotesize
\begin{tabular}{p{0.99\columnwidth} }
 \textcolor{blue}{ {\bf Example 1: 3D Point Cloud + Image} } \\
 \multicolumn{1}{c}{\includegraphics[height=3.5cm]{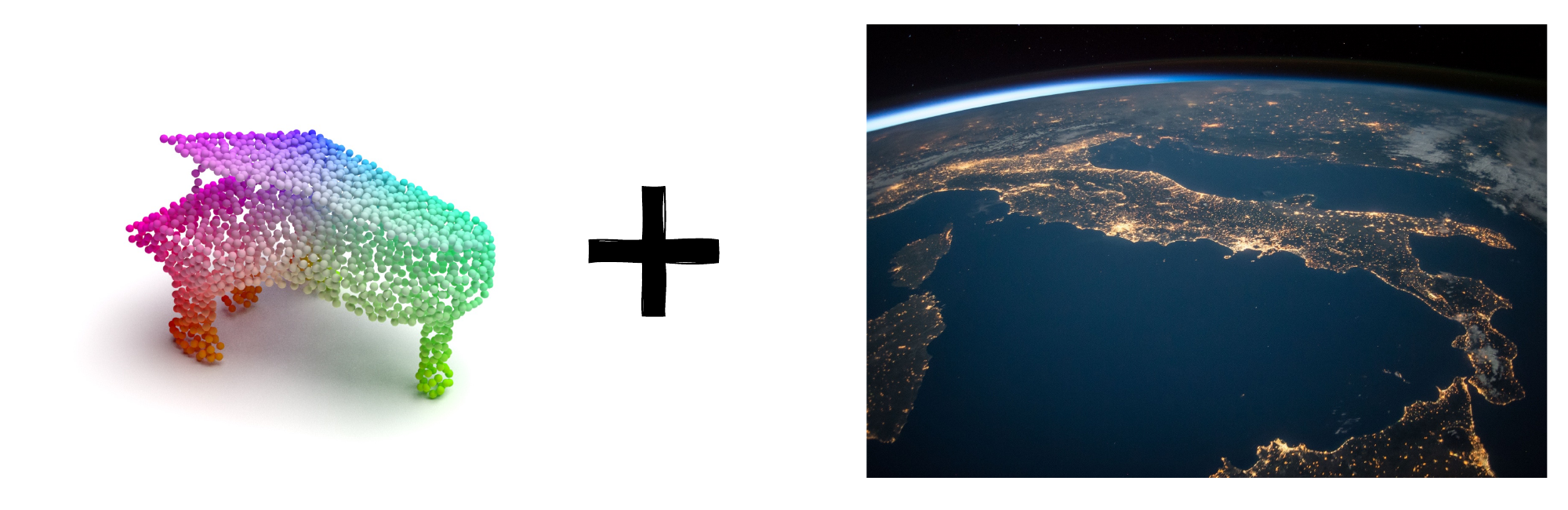}} \\
{\bf Prompt 1:} Describe based on what you see.  \\
{\bf Response 1:} An image of a piano with the earth in the background.  \\ 
{\bf Prompt 2:} What is unusual about this image?  \\
{\bf Response 2:} The unusual aspect of this image is that it depicts a piano floating in space, with the Earth visible in the background. This is an imaginative and creative representation of a musical instrument in an unconventional setting, showcasing its versatility and adaptability to different environments. \\
{\bf Prompt 3:} Imagine the events leading up to this moment and those that follow. Create a story that links them together.  \\
{\bf Response 3:} In the image, we see a grand piano placed on top of the earth, with the moon and stars visible in the background. This setting creates a unique and captivating scene that invites the viewer to imagine the events leading up to this moment and those that follow. One possible storyline could involve a talented pianist who has always dreamed of performing on top of the earth. After years of hard work and dedication, they finally have the opportunity to make their dream come true. As they prepare for their performance, they reflect on the journey that led them to this moment and the challenges they faced along the way. During their performance, they play a beautiful melody that captures the essence of the earth's beauty and the vastness of the universe. The audience is mesmerized by the performance, and the pianist feels a sense of accomplishment and fulfillment as they conclude their performance on top of the earth.  \\ \\
\hrulefill  \\ \\
\textcolor{blue}{ {\bf Example 2: 3D Point Cloud + Image} }  \\
 \multicolumn{1}{c}{\includegraphics[height=3.5cm]{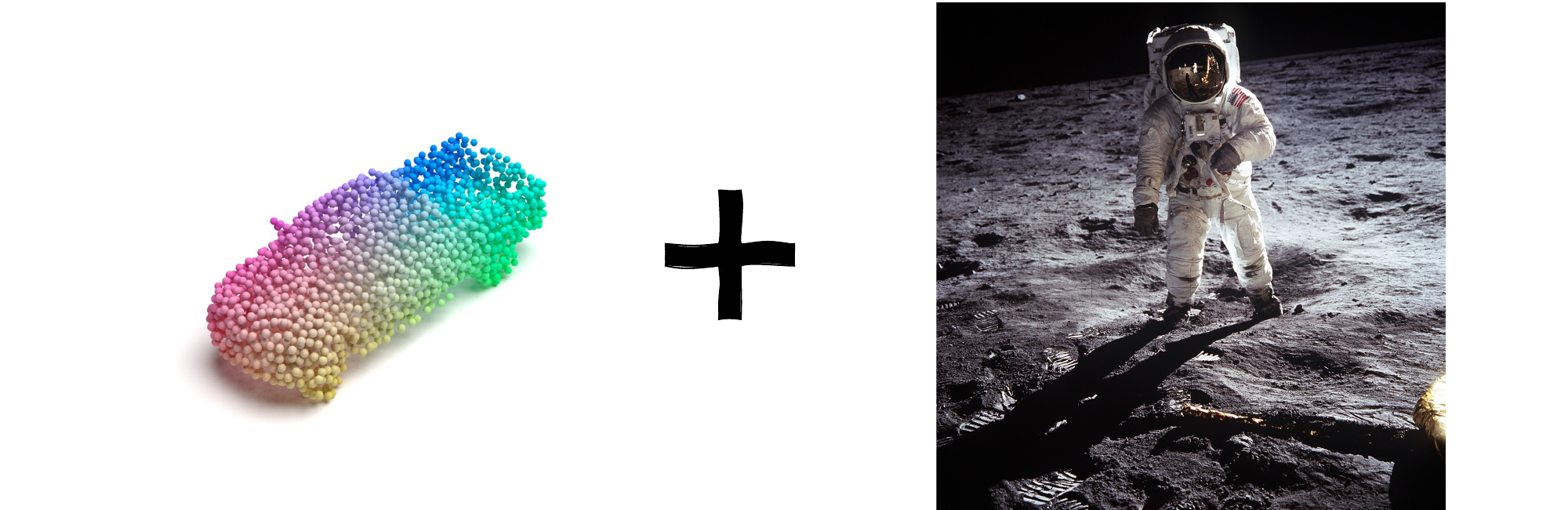}} \\
{\bf Prompt 1:} Imagine you are a narrator looking at this, describe the essence of what you see.  \\
{\bf Response 1:} An astronaut standing next to a pickup truck on the surface of the moon.  \\
{\bf Prompt 2:} What is unusual aboout this image?  \\
{\bf Response 2:} The unusual aspect of this image is that it depicts an astronaut standing on the surface of the moon with a pickup truck in the foreground. This combination of elements is not commonly associated with space exploration or the moon's surface, making it an unconventional and eye-catching image.  \\
{\bf Prompt 3:} Craft a narrative based on the visual elements in the picture. Let your imagination guide the story. \\
{\bf Response 3:} In the image, an astronaut is standing on the moon's surface with a pickup truck in the background. The astronaut is wearing a spacesuit and appears to be exploring the lunar terrain. The pickup truck is parked on the moon's surface, adding a unique touch to the scene. The combination of the astronaut and the pickup truck creates an intriguing narrative that invites the viewer to imagine what might be happening in this futuristic setting.
\end{tabular}
\end{tcolorbox}
\vspace{-2mm}
\end{minipage}
\end{table*}

\begin{table*}[h!]\centering
\begin{minipage}{1.0\columnwidth}\vspace{0mm}
\centering
\caption{Example to illustrate the Instruct-BLIP with three multimodal inputs.}
    \label{tab:example_mm3_insblip}
\begin{tcolorbox}[colback=white,colframe=blue!75!black,title={\bf InstructBLIP w/ \methodname, Three Multimodal Inputs.}]
    \centering
      \footnotesize
\begin{tabular}{p{0.99\columnwidth} }
 \textcolor{blue}{ {\bf Example 1: 3D Point Cloud + Image + Image} } \\
 \multicolumn{1}{c}{\includegraphics[height=2.5cm]{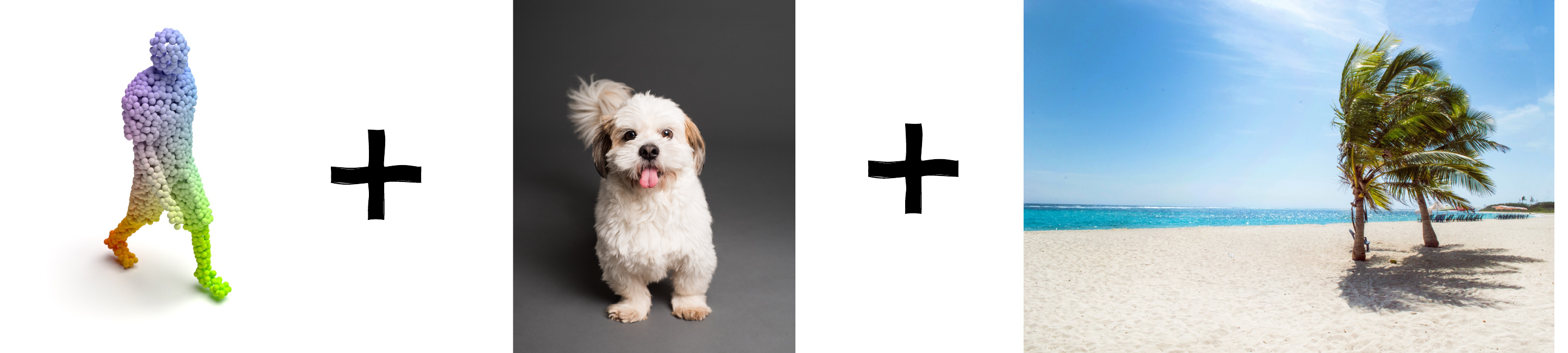}}  \\
{\bf Prompt 1:} Describe the scene based on what you see.  \\
{\bf Response 1:} A man walking a dog on a beach with palm trees in the background.  \\ 
{\bf Prompt 2:} Describe the details of what you see. \\
{\bf Response 2:} The image depicts a man and his dog walking on a sandy beach. The man is holding the dog's leash, while the dog appears to be enjoying its time on the beach. In the background, there is a palm tree visible, adding to the serene atmosphere of the scene. The man and the dog seem to be in good spirits, enjoying each other's company as they stroll along the beach.\\
{\bf Prompt 3:} Imagine the events leading up to this moment and those that follow. Create a story that links them together. \\
{\bf Response 3:} In the story, the man and his dog are walking along the beach, enjoying the sunny day. As they walk, the man notices a piece of paper lying on the sand. Curious, he picks it up and discovers that it's an advertisement for a photography contest. The man decides to enter the contest and submits a photo of himself and his dog on the beach. To his surprise, his photo is selected as one of the winners, and he is invited to attend the award ceremony. At the event, he meets other photographers and learns more about the art of photography. Through this experience, the man gains a new appreciation for photography and decides to pursue it as a hobby. He continues to take photos of his dog and other subjects, honing his skills and capturing unique moments in time. Eventually, his passion for photography leads him to become a professional photographer, and he shares his work with others, inspiring them to appreciate the beauty of the world around them. \\
\\
\hrulefill  \\  
\\ 
\textcolor{blue}{ {\bf Example 2: 3D Point Cloud + Image + Image} }  \\
 \multicolumn{1}{c}{\includegraphics[height=2.5cm]{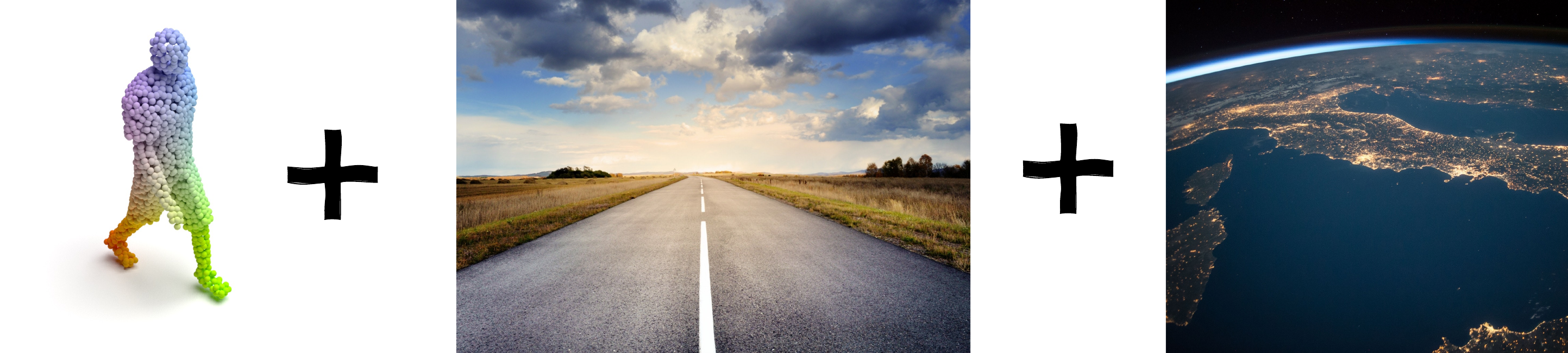}} \\
{\bf Prompt 1:} Describe what you see.  \\
{\bf Response 1:} A man walking down a road with the earth in the background.  \\
{\bf Prompt 2:} What is unusual aboout this image?  \\
{\bf Response 2:} The unusual aspect of this image is that it depicts a man walking on a road, but the background appears to be a space scene with a planet in the distance. This combination of elements creates a surreal and futuristic atmosphere, making the image stand out as unconventional and visually striking. The man's presence on the road and the planet in the background create a sense of contrast between the mundane and the extraordinary, adding an element of surprise and intrigue to the image. \\
{\bf Prompt 3:} Craft a narrative based on the visual elements in the picture. Let your imagination guide the story. \\
{\bf Response 3:} As you look at the image, you see a man standing on a road with a planet in the background. The man is wearing a shirt and shorts, and he appears to be contemplating something as he stands on the road. The planet in the background adds a sense of wonder and exploration to the scene, as if the man is pondering the vastness of the universe and his place within it. Perhaps he is reflecting on his journey so far, or perhaps he is planning his next adventure. In any case, the combination of the man, the road, and the planet creates an intriguing narrative that invites the viewer's imagination to fill in the details.
\end{tabular}
\end{tcolorbox}
\vspace{-2mm}
\end{minipage}
\end{table*}

\clearpage

\section{Related Work}
\textbf{Vision Language Pretraining.} Training images together with linguistic signals, such as words or sentences, has proven to be a powerful approach for tasks like zero-shot recognition and text-to-image retrieval~\cite{faghri2017vse++,frome2013devise,kiros2014unifying,socher2014grounded}. Leveraging language as supervision has also been beneficial for learning robust video representations~\cite{alayrac2020self,miech2020end,miech2019howto100m}. In~\cite{joulin2016learning}, large-scale image datasets with noisy captions were shown to yield strong visual features. Recent advancements in vision-language pretraining, such as CLIP~\cite{openai_clip}, ALIGN~\cite{jia2021align}, and Florence~\cite{yuan2021florence}, have collected extensive image and text pairs and trained models to embed both modalities in a shared space using contrastive learning, leading to impressive zero-shot performance. CoCa~\cite{yu2022coca} introduced an image captioning objective along with contrastive loss for further improvement. Flamingo~\cite{alayrac2022flamingo} handles interleaved images and texts and achieves state-of-the-art results on various few-shot learning benchmarks. LiT~\cite{zhai2022lit} utilizes contrastive training for fine-tuning and observes that freezing image encoders yields the best results. While prior works primarily focus on image and text modalities, \methodname aims to extend the capability of zero-shot recognition to multiple modalities, enabling comprehensive multimodal understanding.

\textbf{Align to CLIP.} 
Pretrained CLIP models have emerged as effective teachers to guide other models due to the robustness of their visual representations~\cite{peng2022beitv2,fang2023eva,wei2022contrastive}. Beyond its role as a teacher, CLIP's joint image and text embedding space has been harnessed for a diverse set of zero-shot tasks, such as segmentation~\cite{li2022languagedriven}, detection~\cite{gu2021open,zhou2022detecting}, 3D shape understanding~\cite{xue2023ulip,xue2023ulip2,liu2023openshape,zhang2022pointclip,zhu2022pointclipv2}, 3D open-vocabulary segmentation~\cite{Peng2023OpenScene}, mesh animation~\cite{youwang2022clipactor}, and more, showcasing the efficacy of this joint embedding space.
In contrast to merely aligning with CLIP's embedding space, \methodname goes a step further by capitalizing on the capabilities of the powerful \ptvit to comprehend and interpret diverse modalities, significantly enhancing its multi-modal understanding capabilities. By enabling a \ptvit to extend beyond its original scope of images, \methodname achieves comprehensive multimodal understanding, paving the way for enhanced zero-shot recognition across a wide range of modalities and tasks.

\textbf{Advances in Multimodal Learning.} Previous research has explored various approaches for jointly training multiple modalities, either in supervised~\cite{girdhar2022omnivore,likhosherstov2021polyvit, gao2020multi} or self-supervised settings~\cite{girdhar2023omnimae,arandjelovic2017look,tian2020contrastive,morgado2021audio, lin2022egocentric}. The effectiveness of image and language pretraining methods, exemplified by CLIP, has sparked investigations into leveraging such approaches to learn deep semantic representations by aligning other modalities with textual inputs. Several methods have adapted CLIP to extract semantically rich representations from different modalities~\cite{luo2022clip4clip,lin2022frozen,fang2021clip2video,xue2022clip-vip, gao2023mist}.
In the context of multimodal zero-shot learning, several approaches have been proposed to align various modalities to CLIP. For instance, AudioCLIP~\cite{guzhov2022audioclip} incorporates audio as an additional modality into the CLIP framework, enabling zero-shot audio classification. ImageBind~\cite{girdhar2023imagebind} aligns six modalities to CLIP using paired image data. ONE-PEACE~\cite{wang2023onepeace} introduces a unified encoder that is pretrained from scratch to align vision, language, and audio. Zhang \emph{et al.}~\cite{zhang2023metatransformer} pretrain a transformer with LAION-2B following the CLIP methodology and freeze the resulting transformer for downstream tasks involving different modalities. In contrast, \methodname takes a unique approach by empowering a \ptvit to effectively comprehend and bind diverse modalities without the need for any manual annotations. 
% This innovative design enhances the multimodal zero-shot understanding capability of \methodname, enabling it to excel in a wide range of tasks across multiple modalities.

% \vspace{-0.2cm}
\section{Conclusion}
% \vspace{-0.2cm}
In this paper,  we present \methodname, a novel approach for advancing omni-modal representation learning by leveraging a \ptvit to comprehend diverse modalities. Our method views a \ptvit as a multi-modal sensor capable of perceiving various modalities, eliminating the need for separate architectures for different modalities and reducing the burden of large-scale data collection. 
In our evaluation of \methodname in 3D shape understanding, we found that \methodname representations effectively capture a wide range of semantic and visual concepts, enabling superior capabilities for open-world 3D shape recognition. By binding 3D shapes to \vit, our model can be integrated with off-the-shelf MLLM, showcasing its potential to enable LLMs to understand and interact with 3D data in a zero-shot manner.
\methodname aims to pave the way towards comprehensive omni-modal representation learning without requiring large-scale data collection, presenting a scalable and efficient solution for leveraging the power of pretrained models in diverse applications. Future work can explore further extensions to additional modalities and emergent abilities to drive advancements in the field of omni-modal representation learning.

\clearpage

\bibliography{egbib}
\bibliographystyle{plain}

\end{document}

%% file: tab/tab_zs_3dcls.tex
\begin{table}[h]
\caption{Zero-shot 3D classification on downstream datasets. Models are pretrained on \dsA ULIP-ShapeNet Triplets, \dsB ULIP2-Objaverse Triplets and \dsC OpenShape Triplets. Measured in accuracy(\%)} \label{tab:zero-shot-all}
\begin{subtable}[h]{0.48\textwidth}
\centering
\caption{Zero-shot 3D of classification on ModelNet40. Models are pretrained on \dsA ULIP-ShapeNet Triplets.}\label{tab:zero-shot-ulip}
\resizebox{\textwidth}{!}{%
\begin{tabular}{c|c|cc}
% \hline
Method               & \begin{tabular}[c]{@{}c@{}}Training \\ Source\end{tabular} & Top1 & Top5 \\ \hline \hline
ULIP-PointNet++(ssg) & \multirow{8}{*}{\dsA}                    & 55.7 & 75.7 \\
ULIP-PointNet++(msg) &                                                            & 58.4 & 78.2 \\
ULIP-PointMLP        &                                                            & 61.5 & 80.7 \\
ULIP-PointBERT       &                                                            & 60.4 & 84.0 \\
 \cellcolor{gray!25}\methodname-OpenAI-B16    &                                                            & \cellcolor{gray!25}61.7 & \cellcolor{gray!25}85.2 \\
 \cellcolor{gray!25}\methodname-Laion2B-B16       &                                                            & \cellcolor{gray!25}65.4 & \cellcolor{gray!25}92.7 \\
 \cellcolor{gray!25}\methodname-OpenAI-L14    &                                                            & \cellcolor{gray!25}63.3 & \cellcolor{gray!25}87.6 \\
 \cellcolor{gray!25}\methodname-Datacomp-L14  &                                                            & \cellcolor{gray!25}\celldouble{70.6}{\upperf{10.2}} & \cellcolor{gray!25}\celldouble{94.4}{\upperf{10.4}} 
% \hline
\end{tabular}
} 
\end{subtable}
\hfill
\begin{subtable}[h]{0.48\textwidth}
\centering
\caption{Zero-shot 3D classification on ModelNet40. Models are pretrained on \dsB ULIP2-Objaverse Triplets.}\label{tab:zero-shot-ulip2}
\resizebox{\textwidth}{!}{%
\begin{tabular}{c|c|cc}
Method              & \begin{tabular}[c]{@{}c@{}}Training \\ Source\end{tabular} & Top1 & Top5 \\ \hline \hline
ULIP2-PointNeXt     & \multirow{6}{*}{\dsB}                                        & 49.0 & 79.7 \\
ULIP2-PointBERT     &                                                            & 70.2 & 87.0 \\
\cellcolor{gray!25}\methodname-OpenAI-B16   &                                                            & \cellcolor{gray!25}73.4 & \cellcolor{gray!25}92.0 \\
\cellcolor{gray!25}\methodname-Laion2B-B16  &                                                            & \cellcolor{gray!25}74.8 & \cellcolor{gray!25}93.8 \\
\cellcolor{gray!25}\methodname-OpenAI-L14   &                                                            & \cellcolor{gray!25}76.1 & \cellcolor{gray!25}93.2 \\
\cellcolor{gray!25}\methodname-Datacomp-L14 &                                                            & \cellcolor{gray!25}\celldouble{80.6}{\upperf{10.4}} & \cellcolor{gray!25}\celldouble{95.8}{\upperf{8.8}} 
\end{tabular}
}
\end{subtable}
\\
\begin{subtable}[h]{\textwidth}
\caption{Zero-shot 3D classification on Objaverse-LVIS, ModelNet40 and ScanObjectNN. Models are pretrained on \dsC OpenShape Triplets. ``\dsC NO LVIS'' denotes exclude all shapes from the Objaverse-LVIS subset. ``\dsC All'' means using all shapes from OpenShape Triplets.}\label{tab:zero-shot-ops}
\resizebox{\textwidth}{!}{%
\begin{tabular}{c|c|ccc|ccc|ccc}
\toprule
\multirow{2}{*}{Method} & \multirow{2}{*}{\begin{tabular}[c]{@{}c@{}}Training\\  Source\end{tabular}}         & \multicolumn{3}{c|}{Objaverse-LVIS} & \multicolumn{3}{c|}{ModelNet40} & \multicolumn{3}{c}{ScanObjectNN} \\ \cmidrule{3-11} 
                        &                                                                                           & Top1       & Top3       & Top5      & Top1      & Top3     & Top5     & Top1      & Top3      & Top5     \\ \midrule
PointCLIP~\cite{zhang2022pointclip}               & \multirow{2}{*}{\begin{tabular}[c]{@{}c@{}}2D inferences, \\ no 3D training\end{tabular}} & 1.9        & 4.1        & 5.8       & 19.3      & 28.6     & 34.8     & 10.5      & 20.8      & 30.6     \\
PointCLIP v2~\cite{zhu2022pointclipv2}            &                                                                                           & 4.7        & 9.5        & 12.9      & 63.6      & 77.9     & 85.0     & 42.1      & 63.3      & 74.5     \\ \midrule
ULIP-PointBERT~\cite{xue2023ulip}          & \multirow{4}{*}{\dsC NO LVIS}                                                         & 21.4       & 38.1       & 46.0      & 71.4      & 84.4     & 89.2     & 46.0      & 66.1      & 76.4     \\
OpenShape-SparseConv~\cite{liu2023openshape}    &                                                                                           & 37.0       & 58.4       & 66.9      & 82.6      & 95.0     & 97.5     & 54.9      & 76.8      & 87.0     \\
OpenShape-PointBERT~\cite{liu2023openshape}     &                                                                                           & 39.1       & 60.8       & 68.9      & 85.3      & 96.2     & 97.4     & 47.2      & 72.4      & 84.7     \\
\cellcolor{gray!25}\methodname-bigG14       &                                                                                           & \cellcolor{gray!25}\celldouble{50.1}{\upperf{11.0}}          & \cellcolor{gray!25}\celldouble{71.3}{\upperf{10.5}}          & \cellcolor{gray!25}\celldouble{78.1}{\upperf{9.2}}         & \cellcolor{gray!25}\celldouble{86.8}{\upperf{1.5}}         & \cellcolor{gray!25}\celldouble{96.8}{\upperf{0.6}}        & \cellcolor{gray!25}\celldouble{97.8}{\upperf{0.3}}        & \cellcolor{gray!25}\celldouble{59.8}{\upperf{3.1}}         & \cellcolor{gray!25}\celldouble{79.3}{\upperf{2.5}}         & \cellcolor{gray!25}\celldouble{87.7}{\upperf{0.7}}        \\ \midrule
ULIP-PointBERT~\cite{xue2023ulip}          & \multirow{4}{*}{\dsC All}                                                            & 26.8       & 44.8       & 52.6      & 75.1      & 88.1     & 93.2     & 51.6      & 72.5      & 82.3     \\
OpenShape-SparseConv~\cite{liu2023openshape}    &                                                                                           & 43.4       & 64.8       & 72.4      & 83.4      & 95.6     & 97.8     & 56.7      & 78.9      & 88.6     \\
OpenShape-PointBERT~\cite{liu2023openshape}     &                                                                                           & 46.8       & 69.1       & 77.0      & 84.4      & 96.5     & 98.0     & 52.2      & 79.7      & 88.7     \\
\cellcolor{gray!25}\methodname-bigG14           &                                                                                           & \cellcolor{gray!25}\celldouble{52.0}{\upperf{5.2}}   & \cellcolor{gray!25}\celldouble{73.3}{\upperf{4.2}}   & \cellcolor{gray!25}\celldouble{79.9}{\upperf{2.9}}     & \cellcolor{gray!25}\celldouble{87.6}{\upperf{3.2}}     & \cellcolor{gray!25}\celldouble{96.6}{\upperf{0.1}}     & \cellcolor{gray!25}\celldouble{98.4}{\upperf{0.4}}     & \cellcolor{gray!25}\celldouble{60.1}{\upperf{3.4}}      & \cellcolor{gray!25}\celldouble{81.0}{\upperf{1.3}}      & \cellcolor{gray!25}\celldouble{90.3}{\upperf{1.6}}   \\ \bottomrule
\end{tabular}
}
\end{subtable}
\end{table}

%% file: fig/graphs/model_dataset_openai.tex
\begin{wrapfigure}{r}{0.35\textwidth}
\centering
        \begin{tikzpicture}
            \begin{axis}[
                xtick={0, 50, 100, 150},
                xticklabels={ , PT.Data\dsA, PT.Data\dsB, },
                ytick={60,75,90},
                grid=both,
                grid style={line width=.1pt, draw=gray!10},
                major grid style={line width=.2pt,draw=gray!50},
                minor tick num=2,
                axis x line*=bottom,
                axis y line*=left,
                height=1.8in,
                width=\linewidth,
                ylabel style= {align=center},
                ylabel={M40@Acc(\%)},
                ylabel near ticks,
                yticklabel style = {font=\small},
                xticklabel style = {font=\footnotesize},
                legend style={cells={align=left}, font={\fontsize{7 pt}{9 pt}\selectfont}, line width=0.25pt},
                legend pos=north west,
            ]
            \addplot[mark=o, mark size=1.8pt, very thick, ADark] plot coordinates {
                (50, 61.7) %
                (100, 73.4) %
            };\label{pgf:openaiB16:img}
            % \addplot[mark=o, very thick, BDark] plot coordinates {
            %     (50, 65.4) %
            %     (100, 74.8) %
            % };\label{pgf:laion2bB16:img}
            \addplot[mark=square*, mark size=1.8pt, very thick, FDark] plot coordinates {
                (50, 63.3) %
                (100, 80.6) %
            };\label{pgf:openaiL14:img}
            % \addplot[mark=square*, mark size=1.8pt, very thick, DDark] plot coordinates {
            %     (50, 70.6) %
            %     (100, 87.4) %
            % };\label{pgf:datacompL14:img}
            \legend{
                OpenAI-B16,
                % Laion2B-B16,
                OpenAI-L14,
                % Datacomp-L14
            }
            \end{axis}
        \end{tikzpicture}
\caption{Zero-shot Classification Performance on ModelNet40. We assess the impact of model scaling using OpenAI-B16 and OpenAI-L14, and analyze the influence of pretraining datasets using \dsA and \dsB.}\label{fig:abla-scaling1}
\end{wrapfigure}
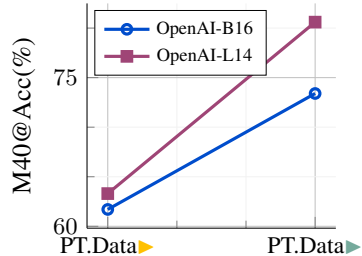

%% file: fig/graphs/scale_cmp_pointbert.tex
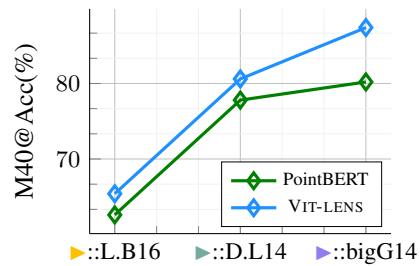
\begin{wrapfigure}{r}{0.4\textwidth}
\centering
        \begin{tikzpicture}
            \begin{axis}[
                xtick={600, 700, 800},
                xticklabels={\dsA::L.B16, \dsB::D.L14, \dsC::bigG14},
                grid=both,
                grid style={line width=.1pt, draw=gray!10},
                major grid style={line width=.2pt,draw=gray!50},
                minor tick num=2,
                axis x line*=bottom,
                axis y line*=left,
                height=1.8in,
                width=\linewidth,
                ylabel style= {align=center},
                ylabel={M40@Acc(\%)},
                ylabel near ticks,
                yticklabel style = {font=\small},
                xticklabel style = {font=\small},
                legend style={cells={align=left}, font={\fontsize{7 pt}{9 pt}\selectfont}, line width=0.25pt},
                legend pos=south east,
            ]
            \addplot[mark=diamond, mark size=3pt, very thick, CDark] plot coordinates {
                (800, 80.2) %
                (700, 77.8) %
                (600,  62.6) %
            };\label{pgf:pointbert_scale:img}
            \addplot[mark=diamond, mark size=3pt, very thick, DDark] plot coordinates {
                (800, 87.4) %
                (700, 80.6) %
                (600,  65.4) %
            };\label{pgf:pointbert_scale:img}
            \legend{
                PointBERT,
                \methodname
            }
            \end{axis}
        \end{tikzpicture}
\caption{Scaling model size and pretraining data: PointBERT vs. \methodname. In experiments, we compare PointBERT and the \methodname's encoder using identical pretraining dataset and CLIP model for alignment.
\methodname exhibits superior zero-shot performance and scalability.}\label{fig:abla-scaling2} \vspace{-1em}
\end{wrapfigure}

%% file: tab/tab_abla.tex
\begin{table}[t]
\caption{Ablation Study. We report top-1 zero-shot accuracy(\%) on ModelNet40.} \label{tab:abla-all}
\begin{subtable}[h]{0.37\textwidth}
\centering
\caption{More comparisons with PointBERT.}\label{tab:abla-3d-pointBERT}
\resizebox{\textwidth}{!}{%
\setlength{\tabcolsep}{1mm}{
\begin{tabular}{l|c|c}
\multicolumn{1}{l|}{PT.D::CLIP Model} & PointBERT & \methodname \\ \Xhline{3\arrayrulewidth}
\dsA :: OpenAI-B16      &    60.2       &     61.7      \\
\dsA :: Laion2B-B16     &    62.6       &     65.4      \\
\dsA :: OpenAI-L14      &    61.2       &     63.3      \\
\dsA :: Datacomp-L14    &    65.4       &     70.6      \\
\dsB :: OpenAI-B16      &    70.6       &     73.4      \\
\dsB :: Laion2B-B16     &    71.7       &     74.8      \\
\dsB :: OpenAI-L14      &    74.1       &     76.1      \\
\dsB :: Datacomp-L14    &    77.8       &     80.6      \\
\dsC :: bigG14          &    84.4       &     87.4          
\end{tabular}
}
}
\end{subtable}
\hfill
\begin{subtable}[h]{0.3\textwidth}
\centering
\caption{Configuration of Perceiver on depth and parameters sharing.}\label{tab:abla-perceiver}
\resizebox{\textwidth}{!}{%
\setlength{\tabcolsep}{1mm}{
\begin{tabular}{cc|c|c}
Depth & \begin{tabular}[c]{@{}c@{}}Share \\ Weights\end{tabular} & \#T.param & Acc@1 \\ \Xhline{3\arrayrulewidth}
2     &        -       &      34.1M    &  64.8     \\
4     &        \xmark     & 67.5M      &  64.2     \\
\rowcolor{lavenderweb}
4     &        \cmark     & 34.1M      &  65.4     \\
6     &         \xmark    & 100.8M     &  65.1     \\
6     &          \cmark   & 34.1M      &  64.0     \\
8     &          \xmark   & 134.2M     &  64.0     \\
8     &          \cmark   & 34.1M      &  64.3     \\
\end{tabular}
}
}
\end{subtable}
\hfill
\begin{subtable}[h]{0.3\textwidth}
\caption{Effect of Perceiver \#latents and ViT position embedding.}\label{tab:abla-hyperparam} % #num latent, use pt pos
\resizebox{\textwidth}{!}{%
\setlength{\tabcolsep}{1mm}{
\begin{tabular}{cc|c|c}
\#latents & ViT.pos & Flops & Acc@1 \\ \Xhline{3\arrayrulewidth}
   128       &   \xmark      &   54.0G   &  65.1   \\
   128       &   \cmark      &   54.0G   &  65.2  \\
   196       &   \xmark      &  75.4G  & 65.1   \\
   \rowcolor{lavenderweb}
   196       &   \cmark      &  75.4G  & 65.4    \\
   256       &   \xmark      &  94.6G  & 65.5   \\
   256       &   \cmark      &  94.6G  & 65.5  \\
   384       &   \xmark      &  136.4G  & 66.2  \\
   384       &    \cmark     &  136.4G  &  66.3    \\
   512       &    \xmark     &  179.5G  &  66.3    \\
   512       &    \cmark     &  179.5G  &  67.4  
\end{tabular}
}
}
\end{subtable}
\\
\begin{subtable}[h]{0.45\textwidth}
\caption{``Point Embedding$\rightarrow$Perceiver'' Performance.}\label{tab:abla-point-perceiver} % #num latent, use pt pos
\resizebox{\textwidth}{!}{%
\setlength{\tabcolsep}{1mm}{
\begin{tabular}{ccc|cc|c}
Depth & \#latents & \begin{tabular}[c]{@{}c@{}}Share \\ Weights\end{tabular} & \#T.param & Flops & Acc@1 \\ \Xhline{3\arrayrulewidth}
 2 & 196       &   -      &  34.1M   & 27.4G  & 62.2  \\
 4 & 196       &   \xmark      &  67.5M  & 40.5G  &  62.4  \\
 8 & 196       &   \xmark      & 134.6M   & 66.7G  & 62.7  \\
 6 & 196       &   \xmark      & 101.2M   & 53.6G  & 61.9  \\
 6 & 196       &   \cmark      & 34.1M   & 53.6G  &  62.3 \\
 6 & 256       &   \xmark      & 101.3M   & 65.6G &  63.5  \\
 6 & 256       &   \cmark      & 34.2M   & 65.6G &   62.7 \\
 6 & 512  &    \xmark     &  101.5M    & 116.6G   &  62.5  \\
 6 & 512  &    \cmark    & 34.4M    &  116.6G &  62.3 \\ \midrule
 \multicolumn{6}{l}{\textit{Default Settings} in \methodname} \\ \Xhline{2\arrayrulewidth}
 \rowcolor{lavenderweb}
 4 & 196  &    \cmark    & 34.1M  &  75.4G  &  65.4
\end{tabular}
}
}
\end{subtable}
\hfill
\begin{subtable}[h]{0.53\textwidth}
\caption{``Point Embedding$\rightarrow$\ptvit'' Performance.}\label{tab:abla-point-vit} % #num latent, use pt pos
\resizebox{\textwidth}{!}{%
\setlength{\tabcolsep}{1mm}{
\begin{tabular}{l|cc|c}
\begin{tabular}[c]{@{}c@{}} Unlocked Components in \vit \end{tabular} & \#T.param & Flops & Acc@1 \\ \Xhline{3\arrayrulewidth}
% \multicolumn{4}{l}{\textit{\# transformer input tokens = 512}} \\ \Xhline{2\arrayrulewidth}
 None   &  7.3K  & 111.4G  &  50.0 \\
 \texttt{[CLS]}     &  7.3K  & 111.4G  &  53.6  \\
  \texttt{[CLS]}, \texttt{Proj}   & 1.1M   & 111.4G &  60.8  \\
  \texttt{[CLS]}, \texttt{Proj}, \texttt{Block.1}, \texttt{Block.2}    & 15.3M     & 111.4G   &  64.8     \\
  \texttt{[CLS]}, \texttt{Proj}, \texttt{Block.11}, \texttt{Block.12}  &  15.3M    & 111.4G   &  64.2    \\
  \texttt{[CLS]}, \texttt{Proj}, \texttt{Block.1} - \texttt{Block.4}    & 29.5M     & 111.4G   &  65.4    \\
  \texttt{[CLS]}, \texttt{Proj}, \texttt{Block.9} - \texttt{Block.12}    & 29.5M     & 111.4G   &  64.7    \\
  \texttt{[CLS]}, \texttt{Proj}, \texttt{Block.1} - \texttt{Block.6}    & 43.7M     & 111.4G   &   66.4   \\
  \texttt{[CLS]}, \texttt{Proj}, \texttt{Block.7} - \texttt{Block.12}    & 43.7M     & 111.4G   &  65.6    \\
  All & 86.6M  &  111.4G     &  67.7     \\ \midrule
  % \multicolumn{4}{l}{\textit{\# transformer input tokens = 196} } \\ \Xhline{2\arrayrulewidth}
  % \texttt{[CLS]}, \texttt{Proj}, \texttt{Block.1}, \texttt{Block.2}    & 15.3M   &  40.4G    &   \\
  % \texttt{[CLS]}, \texttt{Proj}, \texttt{Block.11}, \texttt{Block.12}  &  15.3M    &  40.4G  &    \\
  %  All &   86.3M     &   40.4G          &       \\
  \multicolumn{4}{l}{\textit{Default Settings} in \methodname} \\ \Xhline{2\arrayrulewidth}
  \rowcolor{lavenderweb}
  None(tune PointEmb, Perceiver) & 34.1M & 75.4G &  65.4
\end{tabular}
}
}
\end{subtable}
\end{table}

%% file: tab/tab_mllm_example_plant.tex
\begin{table}[h]
\captionof{table}{Example for \methodname enabling MLLM to understand 3D inputs.}  
\label{tab:visual_example_plant} 
\begin{minipage}{0.99\textwidth}
\centering  
\scalebox{0.88}{
\begin{tabular}{l p{10.5cm} }
\toprule
 \multicolumn{2}{l}{\bf Visual input example, Plant:}  \\
\midrule
&  \includegraphics[height=3.5cm]{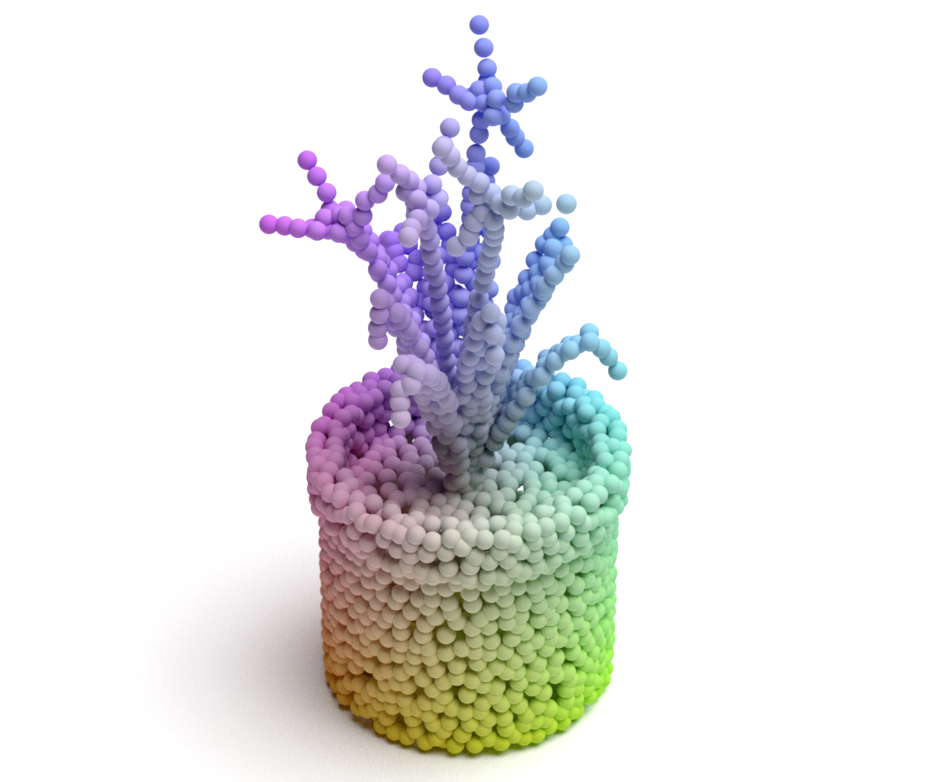} \\
User & What is this object? \\
InstructBLIP w/ \methodname & Potted plant.
\\
User & Describe it in detail. \\
InstructBLIP w/ \methodname & In the image, there is a potted plant \textcolor{blue}{sitting in a ceramic pot} on a table. The plant appears to be a \textcolor{blue}{bamboo-like} species, and it is positioned in such a way that its leaves appear to be growing out of the top of the pot. The plant's placement on the table suggests that it is meant to be a decorative element, adding a touch of greenery to the space. The ceramic pot has a diameter of around 10 centimeters, making it a suitable size for holding a small plant like this one. Overall, the image showcases a simple yet elegant arrangement of a potted plant in a ceramic pot on a table.
\\
\midrule
User & What is this object? \\
InstructBLIP w/ PointBERT &  1000px by 1000px \\
User & Describe this object briefly. \\
InstructBLIP w/ PointBERT & A man in a suit and tie is standing in front of a desk. \\
% User & Describe it in detail. \\
% InstructBLIP w/ PointBERT & The image showcases a man standing in front of a black and white background, wearing a suit and tie. He is holding a briefcase in his right hand and appears to be ready for work or a business meeting. The man's facial expression is serious and focused, conveying a sense of professionalism and dedication to his work. This image captures the essence of a well-dressed, organized, and dedicated individual who is ready to tackle any challenges that may come his way.\\ 
\midrule
CLIPCap-Openshape~\cite{mokady2021clipcap,liu2023openshape} & Flowers in a vase, black and white. \\
\bottomrule
\end{tabular}
} 
\end{minipage}
\end{table}

%% file: tab/tab_mllm_example_piano.tex
\begin{table}[h!]
\captionof{table}{Example for \methodname enabling MLLM to understand 3D inputs.}  
\label{tab:visual_example_piano} 
\begin{minipage}{0.99\textwidth}
\centering  
\scalebox{0.88}{
\begin{tabular}{l p{10.5cm} }
\toprule
 \multicolumn{2}{l}{\bf Visual input example, Piano:}  \\
\midrule
&  \includegraphics[height=3.5cm]{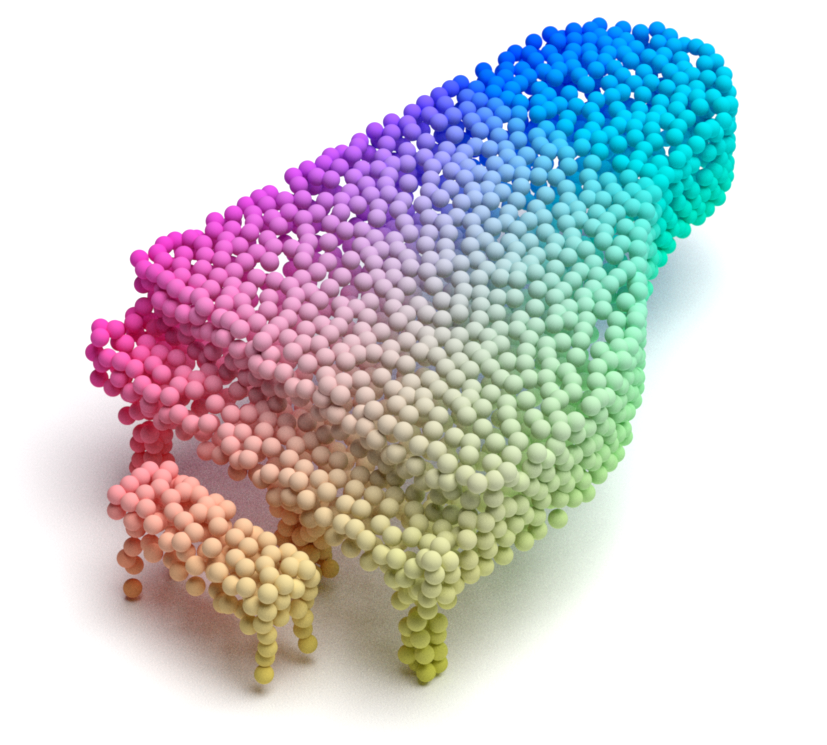} \\

User & What is this object? \\
InstructBLIP w/ \methodname & Piano.\\
User & What is the function of this? \\
InstructBLIP w/ \methodname & Pianos are used to play music.\\
User & Describe it in detail. \\
InstructBLIP w/ \methodname & The object in the image is an upright piano, which is a type of keyboard instrument known for its distinct sound and versatility. The piano is placed on a wooden surface, possibly a floor or a stage, and it appears to be \textcolor{blue}{quite large in size}. Additionally, there is \textcolor{blue}{a chair} placed next to the piano, suggesting that it may be used for seating during performances or lessons. The piano is positioned in the center of the image, making it the focal point of the scene. Overall, the combination of the piano and the chair creates an inviting atmosphere for music enthusiasts or those interested in learning how to play the piano.
\\
\midrule
User & What is this object? \\
InstructBLIP w/ PointBERT &  Desk. \\
User & Describe this object briefly. \\
InstructBLIP w/ PointBERT & A black and white image of an office desk \\
\midrule
CLIPCap-Openshape~\cite{mokady2021clipcap,liu2023openshape} & A white house with a microphone in the middle of a dark room. \\
\bottomrule
\end{tabular}
} 
\end{minipage}
\end{table}

%% file: tab/tab_mllm_example_toilet.tex
\begin{table}[h]
\captionof{table}{Example for \methodname enabling MLLM to understand 3D inputs.}  
\label{tab:visual_example_toilet} 
\begin{minipage}{0.99\textwidth}
\centering  
\scalebox{0.88}{
\begin{tabular}{l p{10.5cm} }
\toprule
 \multicolumn{2}{l}{\bf Visual input example, Toilet:}  \\
\midrule
&  \includegraphics[height=3.5cm]{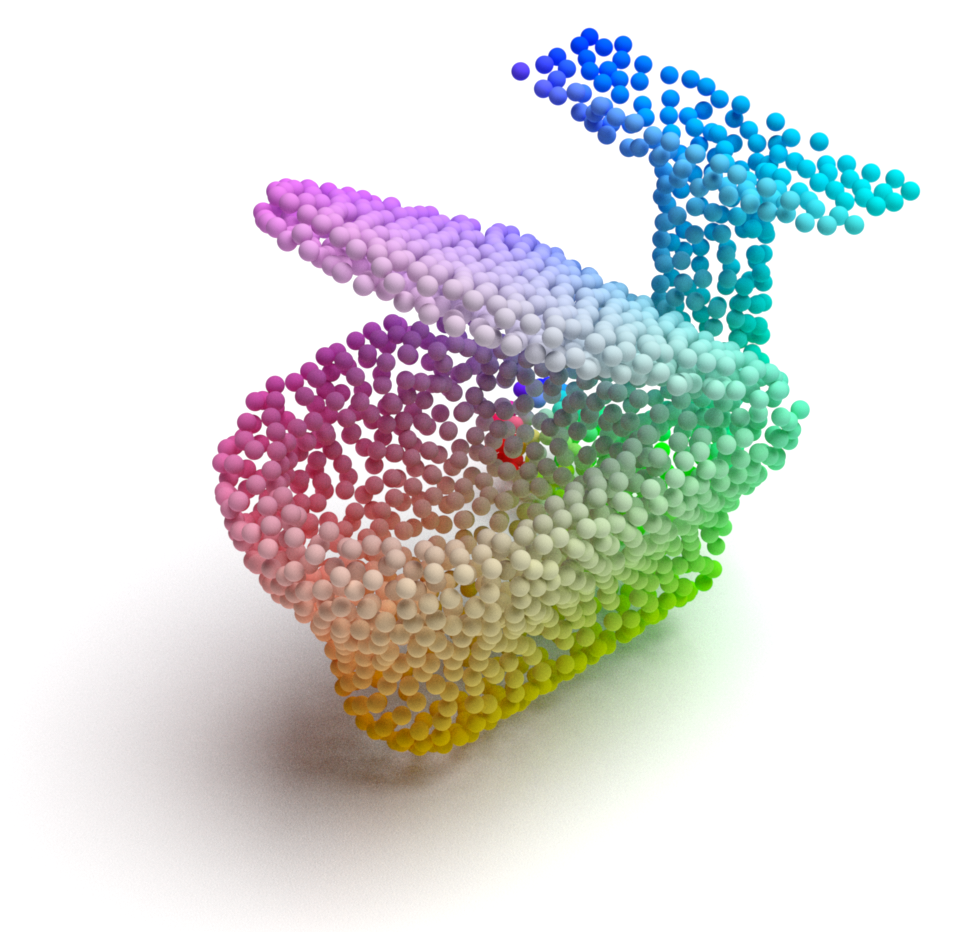} \\

User & What is in the scene? \\
InstructBLIP w/ \methodname & Toilet.\\
User & Describe it in detail. \\
InstructBLIP w/ \methodname & The image features a close-up view of a toilet with its lid open, revealing the inside of the toilet bowl. The toilet appears to be made of porcelain or ceramic material, and it has a seat attached to it. The toilet is positioned on a white background, providing a clear and detailed view of the toilet's design and construction. In addition to the toilet, there is another object in the image, which could be a piece of furniture or a decorative item, but it is not as prominent as the toilet. Overall, the image showcases a functional and well-designed toilet, emphasizing the importance of cleanliness and hygiene in daily life.
\\
\midrule
User & What is in the scene? \\
InstructBLIP w/ PointBERT &  Lamp. \\
User & Describe this object briefly. \\
InstructBLIP w/ PointBERT & An image of a desk with a lamp on it. \\
\midrule
CLIPCap-Openshape~\cite{mokady2021clipcap,liu2023openshape} & The moment a man's hand reaches out to touch a toilet bowl. \\
\bottomrule
\end{tabular}
} 
\end{minipage}
\end{table}